\RequirePackage{fix-cm}
\documentclass[twocolumn]{svjour3}          % twocolumn
\smartqed  % flush right qed marks, e.g. at end of proof
\usepackage{graphicx}
\usepackage{epsfig}
\usepackage{amsmath}
\usepackage{amssymb}
\usepackage{amssymb}
\usepackage{algorithm}
\usepackage{epsfig}
\usepackage{epstopdf}
\usepackage{algpseudocode}
\usepackage{multirow}
\usepackage{subfigure}

\makeatletter
  \newcommand\figcaption{\def\@captype{figure}\caption}
  \newcommand\tabcaption{\def\@captype{table}\caption}
\makeatother

\graphicspath{{Figs/}}

\begin{document}

\title{Unconstrained Still/Video-Based Face Verification with Deep
Convolutional Neural Networks%\thanks{Grants or other notes
}


\author{Jun-Cheng Chen$^*$ \thanks{$^*$ First five authors contributed equally
}     \and
        Rajeev Ranjan$^*$          \and
        Swami Sankaranarayanan$^*$ \and
        Amit Kumar$^*$             \and
        Ching-Hui Chen$^*$         \and
        Vishal M. Patel        \and
        Carlos D. Castillo     \and
        Rama Chellappa         %etc.
}


\institute{Jun-Cheng Chen \at
              A.V. Williams 4455, \\
              University of Maryland, College Park,\\
              MD 20740, USA \\
%              Tel.: +123-45-678910\\
%              Fax: +123-45-678910\\
              \email{pullpull@cs.umd.edu}           %  \\
}

%\institute{F. Author \at
%              first address \\
%              Tel.: +123-45-678910\\
%              Fax: +123-45-678910\\
%              \email{fauthor@example.com}           %  \\
%%             \emph{Present address:} of F. Author  %  if needed

\date{Received: date / Accepted: date}

\maketitle

\begin{abstract}
Over the last five years, methods based on Deep Convolutional Neural
Networks (DCNNs) have shown impressive performance improvements for
object detection and recognition problems. This has been made
possible due to the availability of large annotated datasets, a
better understanding of the non-linear mapping between input images
and class labels as well as the affordability of GPUs. In this
paper, we present the design details of a deep learning system for
unconstrained face recognition, including modules for face
detection, association, alignment and face verification. The
quantitative performance evaluation is conducted using the IARPA
Janus Benchmark A (IJB-A), the JANUS Challenge Set 2 (JANUS CS2),
and the LFW dataset. The IJB-A dataset includes real-world
unconstrained faces of 500 subjects with significant pose and
illumination variations which are much harder than the Labeled Faces
in the Wild (LFW) and Youtube Face (YTF) datasets. JANUS CS2 is the
extended version of IJB-A which contains not only all the
images/frames of IJB-A but also includes the original videos. Some
open issues regarding DCNNs for face verification problems are then
discussed.

\end{abstract}

\section{Introduction}\label{intro}

Face verification is a challenging problem in computer vision and
has been actively researched for over two decades
\cite{zhao_face_2003}. In face verification, given two videos or
images, the objective is to determine whether they belong to the
same person. Many algorithms have been shown to work well on images
and videos that are collected in controlled settings. However, the
performance of these algorithms often degrades significantly on
images that have large variations in pose, illumination, expression,
aging, and occlusion. In addition, for an automated face
verification system to be effective, it also needs to handle errors
that are introduced by algorithms for automatic face detection, face
association, and facial landmark detection.

Existing methods have focused on learning robust and discriminative
representations from face images and videos. One approach is to
extract an over-complete and high-dimensional feature representation
followed by a learned metric to project the feature vector onto a
low-dimensional space and then compute the similarity scores. For
example, high-dimensional multi-scale local binary pattern (LBP)
\cite{chen_blessing_2012} features extracted from local patches
around facial landmarks and Fisher vector (FV)
\cite{simonyan_fisher_2013,Chan_FV_BTAS_2015} features have been
shown to be effective for face recognition. Despite significant
progress, the performance of these systems has not been adequate for
deployment. However, given the availability of millions of annotated
data, faster GPUs and a better understanding of the nonlinearities,
DCNNs are providing much better performance on tasks such as object
recognition~\cite{krizhevsky_imagenet_2012,szegedy_going_2014},
object/face detection~\cite{girshick_rich_2014,ranjan_deep_2015},
face
verification/recognition~\cite{schroff_facenet_2015,parkhi_deep_2015}.
It has been shown that DCNN models can not only characterize large
data variations but also learn a compact and discriminative
representation when the size of training data is sufficiently large.
In addition, it can be generalized to other vision tasks by
fine-tuning the pre-trained model on the new
task~\cite{donahue_decaf_2013}.

In this paper, we present an automated face verification system. Due
to the robustness of DCNNs, we build each component of our system
based on separate DCNN models. Modules for detection and face
alignment use the DCNN architecture proposed
in~\cite{krizhevsky_imagenet_2012}. For face verification, we train
two DCNN models trained using the
CASIA-WebFace~\cite{yi_learning_2014} dataset. Finally, we compare
the performance of our approach with many face matchers on the IJB-A
dataset which are being carried out or have been recently
reported~\cite{nist_ijba_2016}\footnote{While this paper was under
review, several recent works have also reported improved numbers on
the IJB-A dataset~\cite{ranjan2016all} and its successive version
Janus Challenge Set 3 (CS3)~\cite{navaneeth2017fusion}. We refer the
interested readers to these works for more details.}The proposed
system is fully automatic. Although the IJB-A dataset contains
significant variations in pose, illumination, expression, resolution
and occlusion which are much harder than the Labeled Faces in the
Wild (LFW) datasets, we present verification results for the LFW
dataset too.\\
\indent The system described in this paper, which integrates
DCNN-based face detection~\cite{ranjan_deep_2015} and fiducial point
detection~\cite{kumar_face_2016} modules differs from its
predecessor~\cite{chen2015end} in the following ways: (1) uses more
robust features from two networks which take faces as input with
different resolutions (Section~\ref{method:representation}) are used
and (2) employs a more efficient metric learning method~\cite{tse}
which uses inner-products based constraints between triplets to
optimize for the embedding matrix as opposed to norm-based
constraints used in other methods (Section~\ref{method:metric}). In
the experimental section, we also demonstrate the improvement due to
media-sensitive pooling and the fusion of two networks.

\indent The rest of the paper is organized as follows. We briefly
review closely related works in Section \ref{sec:related}. In
Section \ref{sec:method}, we present the design details of a deep
learning system for unconstrained face verification and recognition,
including face detection, face association, face alignment, and face
verification. Experimental results using IJB-A, CS2, and LFW
datasets are presented in Section \ref{sec:exp}. Some open issues
regarding the use of DCNNs for face recognition/verification
problems are discussed in Section \ref{sec:open}. Finally, we
conclude the paper in Section \ref{sec:conc} with a brief summary
and discussion.
\begin{figure*}[htp!]
\begin{center}
 \includegraphics[width=6.5in]{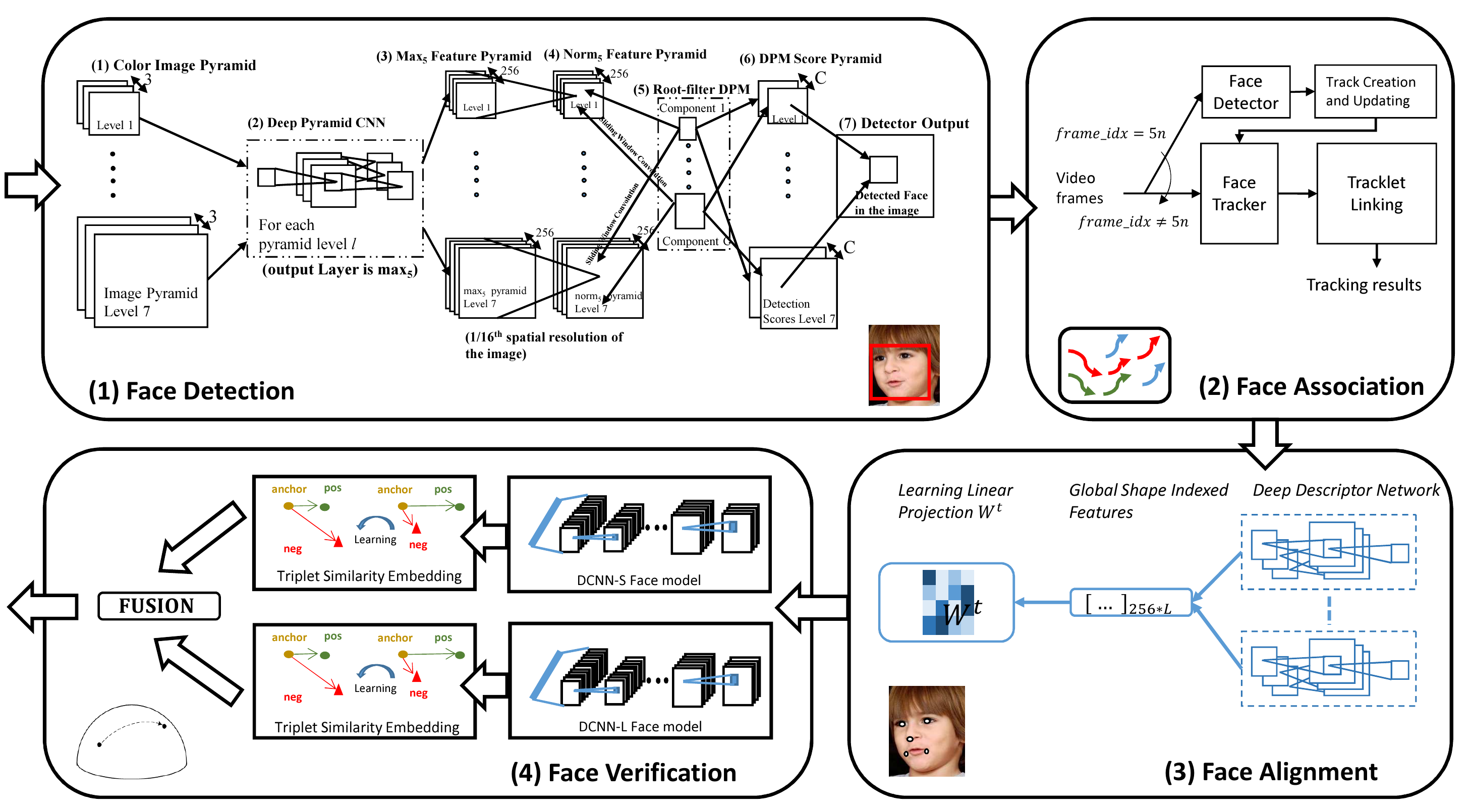}
\end{center}
  \caption{An overview of the proposed DCNN-based face verification system. %\cite{Janus_ICCV_2015}.
  }
  \label{fig:system_overview}
\end{figure*}

\section{Related Work} \label{sec:related}

A typical face verification system consists of the following
components: (1) face detection and (2) face association across video
frames, (3) facial landmark detection to align faces, and (4) face
verification to verify a subject's identity. Due to the large number
of published papers in the literature, we briefly review some
relevant works for each component.

\subsection{Face Detection}
The face detection method introduced by Viola and Jones
\cite{viola_robust_2004} is based on cascaded classifiers built
using Haar wavelet features. Since then, a variety of sophisticated
cascade-based face detectors such as Joint Cascade
\cite{JointCascade_LI_ECCV2014}, SURF Cascade \cite{6619289} and
CascadeCNN \cite{CascadeCNN_CVPR2015} have demonstrated improved
performance. Zhu \emph{et al.} \cite{zhu2012face} improved the
performance of face detection algorithm using the deformable part
model (DPM) approach, which treats each facial landmark as a part
and uses HOG features to simultaneously perform face detection, pose
estimation, and landmark localization. A recent face detector,
Headhunter \cite{HeadHunter_Mathias_ECCV2014}, demonstrated
competitive performance using a simple DPM. However, the key
challenge in unconstrained face detection is that features like Haar
wavelets and HOG do not capture the salient facial information at
different poses and illumination conditions. To overcome these
limitations, several deep CNN-based face detection methods have been
proposed in the literature such as Faceness
\cite{faceness_ICCV2015}, DDFD \cite{DDFD_ICMR2015} and CascadeCNN
\cite{CascadeCNN_CVPR2015}. It has been shown in
\cite{donahue_decaf_2013} that a deep CNN pre-trained with the
Imagenet dataset can be used as a meaningful feature extractor for
various vision tasks. The method based on Regions with CNN (R-CNN)
\cite{girshick_fastrcnn_15} computes region-based deep features and
attains state-of-art face detection performance. In addition, since
the deep pyramid \cite{girshick_deformable_2014} removes the
fixed-scale input dependency in deep CNNs, it is attractive to be
integrated with the DPM approach to further improve the detection
accuracy across scale \cite{ranjan_deep_2015}.

\subsection{Face Association}
Video-based face verification systems \cite{Chen2012_MOT} requires
consistently-tracked faces to capture diverse pose and
spatial-temporal information for analysis. In addition, there is
usually more than one person present in the videos, and thus
multiple face images from different individuals should be correctly
associated across the video frames. Several recent techniques have
tracked multiple objects by modeling the motion context
\cite{Yoon2015MOT}, track management \cite{Duffner2013MOT}, and
guided tracking using the confidence map of the detector
\cite{Breitenstein2009MOT}. Multi-object tracking methods based on
tracklet linking \cite{Huang2008MOT,Roth2012,Bae2014} usually rely
on the Hungarian algorithm \cite{Ahuja1993} to optimally assign the
detected bounding boxes to existing tracklets. Roth \emph{et al.}
\cite{Roth2012} adapted the framework of multi-object tracking
methods based on tracklet linking approach to track multiple faces;
Several face-specific metrics and constraints have been introduced
to enhance the reliability of face tracker. A recent study
\cite{Comaschi2015MOT} proposed to manage the tracks generated by a
continuous face detector without relying on long-term observations.
In unconstrained scenarios, the camera can undergo abrupt movements,
which makes persistent tracking a challenging task. Du \emph{et al.}
proposed a conditional random field (CRF) framework for face
association in two consecutive frames by utilizing the affinity of
facial features, location, motion, and clothing appearance
\cite{Du2012}. Our face association method utilizes the KLT tracker
to track a face initiated from the face detection. We continuously
update the face tracking results for every fifth frame using the
detected faces. The tracklet linking~\cite{Bae2014} is utilized to
link the fragmented tracklet. We present a robust face association
method based on existing works in
\cite{Everingham2009MOT,Bae2014,Shi1994}. In addition, recently
developed object
trackers~\cite{babenko2009visual,henriques2015high,kalal2012tracking}
and face trackers~\cite{wang2008robust,lui2010adaptive} can be
integrated to potentially improve the robustness of face association
method. More details are presented in
Section~\ref{method:face_association}.

\subsection{Facial Landmark Detection}
Facial landmark detection is an important component to align the
faces into canonical coordinates and to improve the performance of
verification algorithms. Pioneering works such as Active Appearance
Models (AAM)~\cite{cootes_active_2001} and Active Shape Models
(ASM)~\cite{cootes_active_1995} are built using the PCA constraints
on appearance and shape. In \cite{cristinacce_feature_2006},
Cristinacce \emph{et al.} generalized the ASM model to a Constrained
Local Model (CLM), in which every landmark has a shape constrained
descriptor to capture the appearance. Zhu \emph{et
al.}~\cite{zhu2012face} used a part-based model for face detection,
pose estimation and landmark localization assuming the face shape to
be a tree structure. Asthana \emph{et
al.}~\cite{asthana_robust_2013} combined the discriminative response
map fitting with CLM. In addition, Cao \emph{et.
al.}~\cite{cao2014face} followed the cascaded pose regression (CPR)
proposed by $Doll\acute{a}r$ \emph{et.
al.}~\cite{dollar2010cascaded}: feature extraction followed by a
regression stage. However, unlike CPR which uses pixel difference as
features, it trains a random forest based on local binary patterns.
In general, these methods learn a model that directly maps the image
appearance to the target output. Nevertheless, the performance of
these methods depends on the robustness of local descriptors.
In~\cite{krizhevsky_imagenet_2012}, the deep features are shown to
be robust to different challenging variations. Sun \emph{et
al.}~\cite{sun_deep_2013} proposed a cascade of carefully designed
CNNs, in which at each level, outputs of multiple networks are fused
for landmark estimation and achieve good performance.
Unlike~\cite{sun_deep_2013}, we use a single CNN, carefully designed
to provide a unique key-point descriptor and achieve better
performance. Besides using a 2D transformation for face alignment,
Hassner \emph{et al.}~\cite{hassner2015effective} proposed an
effective method to frontalize faces with the help of generic 3D
face model. However, the effectiveness of the method also highly
relies on the quality of the detected facial landmarks (\emph{i.e.},
the method usually introduces undesirable artifacts when the quality
of facial landmarks is poor).

\subsection{Feature Representation for Face Recognition}
%\noindent \textbf{Feature Learning:}
Learning invariant and discriminative feature representations is a
critical step in designing a face verification system. Ahonen
\emph{et al.}~\cite{ahonen_face_2006} showed that the Local Binary
Pattern (LBP) is effective for face recognition. Chen \emph{et
al.}~\cite{chen_blessing_2012} demonstrated good results for face
verification using high-dimensional multi-scale LBP features
extracted from patches extracted around facial landmarks. However,
recent advances in deep learning methods have shown that compact and
discriminative representations can be learned using a DCNN trained
with very large datasets. Taigman \emph{et
al.}~\cite{taigman_deepface_2014} built a DCNN model on the
frontalized faces generated with a general 3D shape model from a
large-scale face dataset and achieved better performance than many
traditional methods. Sun \emph{et al.}~\cite{sun_deeply_2014}
achieved results that surpass human performance for face
verification on the LFW dataset using an ensemble of 25 simple DCNN
with fewer layers trained on weakly aligned face images from a much
smaller dataset than \cite{taigman_deepface_2014}. Schroff \emph{et
al.} \cite{schroff_facenet_2015} adapted a state-of-the-art object
recognition network to face recognition and trained it using a
large-scale unaligned private face dataset with triplet loss. Parkhi
\emph{et al.} \cite{parkhi_deep_2015} trained a very deep
convolutional network based on VGGNet for face verification and
demonstrated impressive results. These studies essentially
demonstrate the effectiveness of the DCNN model for feature learning
and detection/recognition/verification problems.

\subsection{Metric Learning}
Learning a similarity measure from data is the other key component
for improving the performance of a face verification system. Many
approaches have been proposed in the literature that essentially
exploit label information from face images or face pairs. For
instance, Weinberger \emph{et al.}~\cite{weinberger_distance_2005}
used the Large Margin Nearest Neighbor (LMNN) metric which enforces
the large margin constraint among all triplets of labeled training
data. Guillaumin \emph{et al.}~\cite{guillaumin2009you} proposed two
robust distance measures: Logistic Discriminant-based Metric
Learning (LDML) and Marginalized kNN (MkNN). The LDML method learns
a distance by performing a logistic discriminant analysis on a set
of labeled image pairs and the MkNN method marginalizes a
k-nearest-neighbor classifier to both images of the given test pair
using a set of labeled training images. Mignon \emph{et
al.}~\cite{mignon2012pcca} proposed an algorithm for learning
distance metrics from sparse pairwise similarity/dissimilarity
constraints in high dimensional input space. The method exhibits
good generalization properties when projecting the features from a
high-dimensional space to a low-dimensional one. Nguyen \emph{et
al.}~\cite{nguyen2010cosine} used an efficient and simple metric
learning method based on the cosine similarity measure instead of
the widely adopted Euclidean distance. Taigman \emph{et al.}
\cite{taigman_multiple_2009} employed the Mahalanobis distance using
the Information Theoretic Metric Learning (ITML) method
\cite{davis_information_2007}. Chen \emph{et al.}
\cite{chen_bayesian_2012} used a joint Bayesian approach for face
verification which models the joint distribution of a pair of face
images and uses the ratio of between-class and within-class
probabilities as the similarity measure. Hu \emph{et al.}
\cite{hu_discriminative_2014} learned a discriminative metric within
the deep neural network framework. Schroff \emph{et al.}
\cite{schroff_facenet_2015} and Parkhi \emph{et al.}
\cite{parkhi_deep_2015} optimized the DCNN parameters based on the
triplet loss which directly embeds the DCNN features into a
discriminative subspace and presented promising results for face
verification.

\section{Proposed System} \label{sec:method}

The proposed system is a complete pipeline for performing
\emph{automatic} face verification. We first perform face detection
to localize faces in each image and video frame. Then, we associate
the detected faces with the common identity across video frames and
align the faces into canonical coordinates using the detected
landmarks. Finally, we perform face verification to compute the
similarity between a pair of images/videos. The system is
illustrated in Figure~\ref{fig:system_overview}. The details of each
component are presented in the following sections.

\subsection{Face Detection}\label{face_detection}
All the faces in the images/video frames are detected using a
DCNN-based face detector, called the Deep Pyramid Deformable Parts
Model for Face Detection (DP2MFD) \cite{ranjan_deep_2015}, which
consists of two modules. The first module generates a seven level
normalized deep feature pyramid for any input image of arbitrary
size, as illustrated in the first part of Figure
\ref{fig:system_overview}. The architecture of
Alexnet~\cite{krizhevsky_imagenet_2012} is adopted for extracting
the deep features. This image pyramid network generates a pyramid of
256 feature maps at the fifth convolution layer (conv$_5$). A 3
$\times$ 3 max filter is applied to the feature pyramid at a stride
of one to obtain the max$_5$ layer. Typically, the activation
magnitude for a face region decreases with the size of the pyramid
level. As a result, a large face detected by a fixed-size sliding
window at a lower pyramid level will have a high detection score
compared to a small face getting detected at a higher pyramid level.
In order to reduce this bias to face size, we apply a z-score
normalization step on the max$_5$ features at each level. For a
256-dimensional feature vector $x_{i,j,k}$ at the pyramid level
\emph{i} and location $(j, k)$, the normalized feature $x_{i,j,k}$
is computed as:

\begin{equation}
 x_{i,j,k} = \frac{x_{i,j,k} - \mu_i}{\sigma_i},
\end{equation}

\noindent where $\mu_i$ is the mean feature vector, and $\sigma_i$ is the
standard deviation for the pyramid level \emph{i}. We refer to the
normalized max$_5$ features as $norm_5$. Then, the fixed-length
features from each location in the pyramid are extracted using the
sliding window approach.

\begin{figure*}[tb]
\begin{center}
\subfigure[]{
 \includegraphics[width=2in]{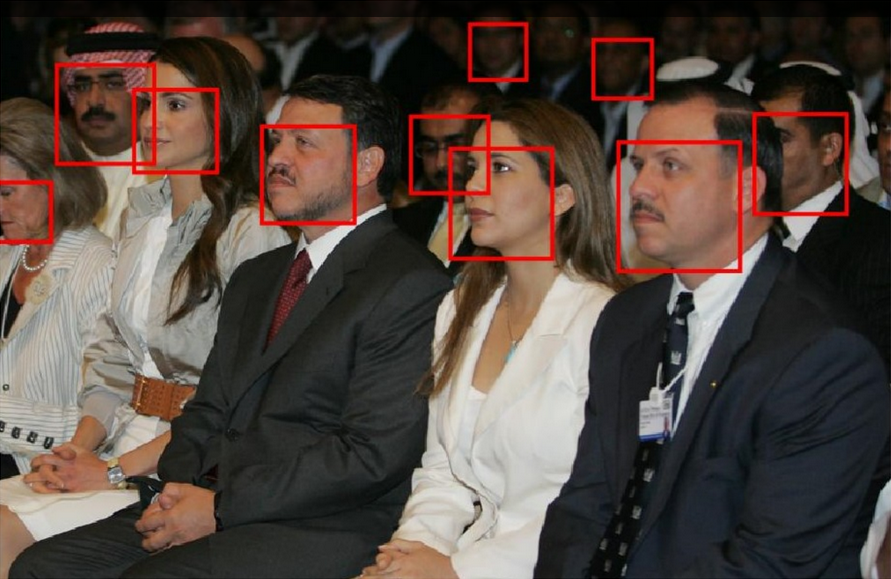}
} \subfigure[]{
 \includegraphics[width=2in]{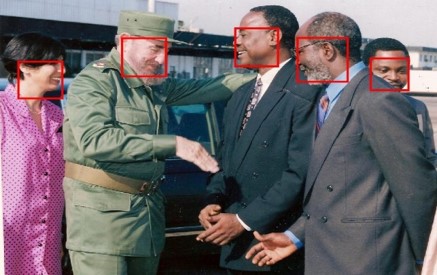}
} \subfigure[]{
 \includegraphics[width=2in]{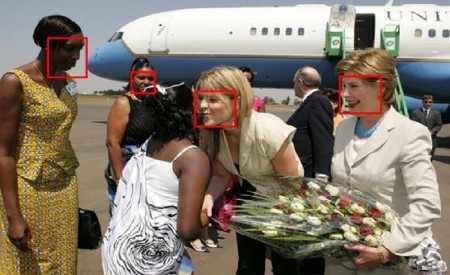}
}
\end{center}
  \caption{Sample detection results on an IJB-A image using the deep pyramid method.}
  \label{exp:face_detection}
\end{figure*}

The second module is a linear SVM, which takes these features as
inputs to classify each location as face or non-face, based on their
scores. A root-only DPM is trained on the norm$_5$ feature pyramid
using a linear SVM. In addition, the deep pyramid features are
robust to not only pose and illumination variations but also to
different scales. The DP2MFD algorithm works well in unconstrained
settings as shown in Figure \ref{exp:face_detection}. We also
present the face detection performance results under the face
detection protocol of the IJB-A dataset in Section ~\ref{sec:exp}.

\subsection{Face Association}\label{method:face_association}

\begin{figure*}[tb]
\begin{center}
 \includegraphics[height=1.7in]{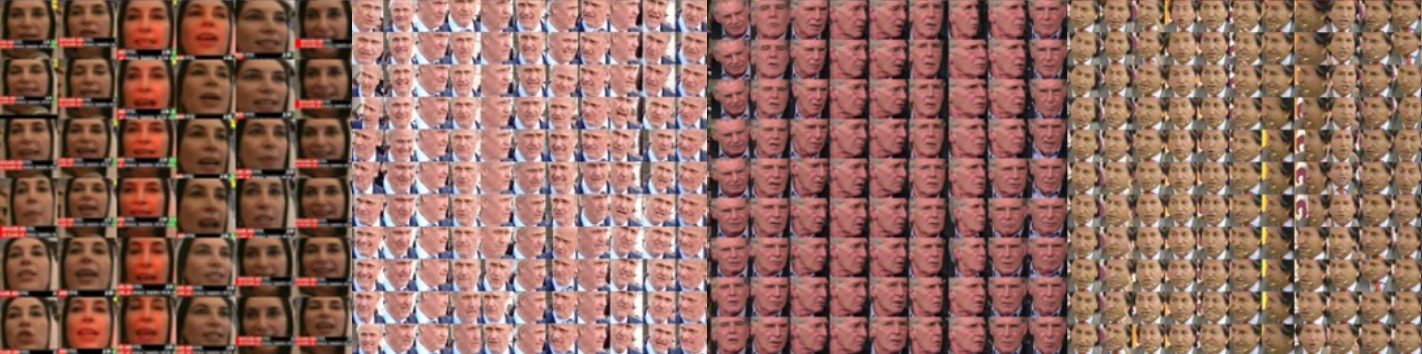}
\end{center}
  \caption{Sample results of our face association method for videos of JANUS CS2 which is
  the extension dataset of IJB-A.}
  \label{exp:face_association}
\end{figure*}

Because there are multiple subjects appearing in the frames of each
video of the IJB-A dataset, performing face association to assign
each face to its corresponding subject is an important step to pick
the correct subject for face verification. Thus, once the faces in
the images and video frames are detected, we track multiple faces by
integrating results from the face detector, face tracker,
 and a tracklet linking step. The second part of Figure \ref{fig:system_overview} shows the block diagram of the multiple
 face tracking system. We apply the face detection algorithm in every fifth frame using the face
 detection method presented in Section \ref{face_detection}. The detected bounding box is considered as a novel
 detection if it does not have an overlap ratio with any bounding box in the previous frames larger than $\gamma$. The overlap ratio of a detected bounding box $\mathbf{b}_{d}$ and
 a bounding box $\mathbf{b}_{tr}$ in the previous frames is defined as
\begin{equation}
\begin{aligned}
s(\mathbf{b}_{d}, \mathbf{b}_{tr}) = \frac{area(\mathbf{b}_{d} \cap
\mathbf{b}_{tr})}{area( \mathbf{b}_{tr})}.
\end{aligned}
\end{equation}
We empirically set the overlap threshold $\gamma$ to $0.2$. A face
tracker is created from a detection bounding box that is treated as
a novel detection. We set the face detection confidence threshold to
-1.0 to select bounding boxes of face detection of high confidence.
For face tracking, we use the Kanade-Lucas-Tomasi (KLT) feature
tracker \cite{Shi1994} to track the faces between two consecutive
frames. To avoid the potential drift of trackers, we update the
bounding boxes of the tracker by those provided by the face detector
in every fifth frame. The detection bounding box $\mathbf{b}_{d}$
replaces the tracking bounding boxes $\mathbf{b}_{tr}$ of a tracklet
in the previous frame if $s(\mathbf{b}_{d}, \mathbf{b}_{tr}) \leq
\gamma$. A face tracker is terminated if there is no corresponding
face detection overlapping with it for more than $t$ frames. We set
$t$ to 4 based on empirical
grounds.\\
\indent In order to handle the fragmented face tracks resulting from
occlusions or unreliable face detection, we use the tracklet linking
method proposed by \cite{Bae2014} to associate the bounding boxes in
the current frames with tracklets in the previous frames. The
tracklet linking method consists of two stages. The first stage is
to associate the bounding boxes provided by the tracker or the
detector in the current frame with the existing tracklet in previous
frames. This stage consists of local and global associations. The
local association step associates the bounding boxes with the set of
tracklets, having high confidence. The global step associates the
remaining bounding boxes with the set of tracklets of low
confidence. The second stage is to update the confidence of the
tracklets, which will be used for determining the tracklets for
local or global association in the first stage. We show sample face
association results for some videos from the CS2 dataset in Figure
\ref{exp:face_association}.

\subsection{Facial Landmark Detection}\label{alignment}
\begin{figure}[tb]
\begin{center}
 \includegraphics[width=0.5\textwidth]{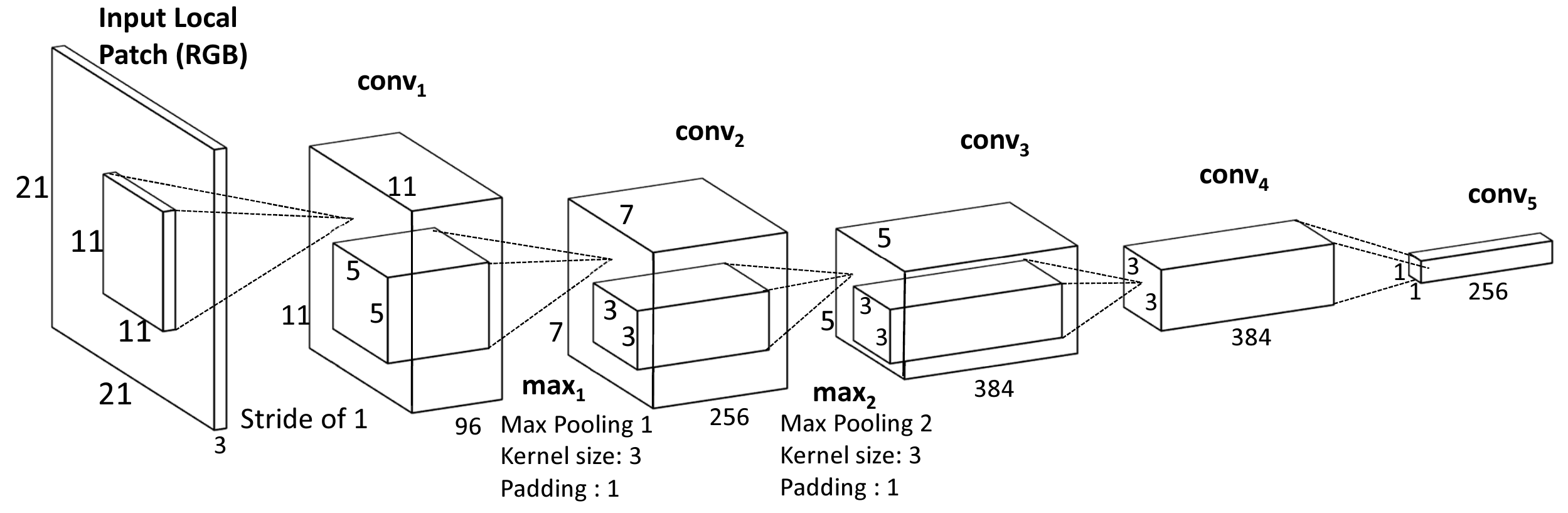}
\end{center}
  \caption{The DCNN architecture used to extract the local descriptors for the facial landmark detection task \cite{kumar_face_2016}.}
  \label{method:face_landmark_dcnn}
\end{figure}

Once the faces are detected or associated, we perform facial
landmark detection for face alignment. The DCNN-based facial
landmark detection algorithm module, local deep descriptor
regression (LDDR) \cite{kumar_face_2016}, works in two stages. We
model the task as a regression problem, where beginning with the
initial mean shape, the target shape is reached through regression.
The first step is to perform feature extraction of a patch around a
point of the shape followed by linear regression as described in
\cite{6909614,cao2014face}. Given a face image $I$ and the initial
shape $S^{0}$, the regressor computes the shape increment $\Delta S$
from the deep descriptors and updates the face shape using
(\ref{update_eq}).
\begin{equation}
\label{update_eq}
% \textit{$S^{t} $} = \textit{$S^{t-1}$} + \textit{$W^t$}\Phi^{t}(\textit{I},\textit{$S^{t-1}$}).
 S^{t}=S^{t-1}+W^{t}\Phi^{t}(I,S^{t-1})
\end{equation}
The CNN features (represented as $\Phi$ in \ref{update_eq})
carefully designed with the proper number of strides and pooling
(refer to Table~\ref{strides} for more details), are used as
features to perform regression. We use the same CNN architecture as
Alexnet \cite{krizhevsky_imagenet_2012} with the pretrained weights
for the ImageNet dataset as shown in Figure
\ref{method:face_landmark_dcnn}. Then, we further fine-tuned it with
AFLW~\cite{tugraz:icg:lrs:koestinger11b} dataset for face detection
task. The fine-tuning step helps the network to learn features
specific to faces. Furthermore, we adopt the cascade regression, in
which the output generated by the first stage is used as an input
for the next stage. The number of stages is fixed at 5 in our
system. The patches selected for feature extraction are reduced
subsequently in later stages to improve the localization of facial
landmarks. After the facial landmark detection is completed, each
face is aligned into the canonical coordinate using the similarity
transform and seven landmark points (\emph{i.e.}, two left eye
corners, two right eye corners, nose tip, and two mouth corners).
\begin{table}[htp!]
\begin{center}
\resizebox{\linewidth}{!}{%
\begin{tabular}{|p{1.1cm}|p{1.6cm}|p{0.7cm}|p{.7cm}|p{.7cm}|p{.7cm}|}
\hline
 \centering Stage 1 &  \centering Input Size (pixels) &  \centering conv1 &  \centering max1 & \centering conv2 &  max2\\
\hline\hline
Stage 1        &\centering $92 \times 92$       &\centering 4    &\centering 2    &\centering 1     & 1 \\
Stage 2          &\centering $68 \times 68$          &\centering 3    &\centering 2     &\centering 1     & 1 \\
Stage 3           &\centering $42 \times 42$              &\centering 2    &\centering 1     &\centering 1     & 2 \\
Stage 4            & \centering $21 \times 21$                 &\centering 1    &\centering 1     &\centering 1    & 1 \\
\hline
\end{tabular}}
\vskip 4pt \caption{Input size and the number of strides in conv1,
max1, conv2 and max2 layers for 4 stages of
regression~\cite{kumar_face_2016}.} \label{strides}
\end{center}
\end{table}

\subsection{Deep Convolutional Face Representation}
\label{method:representation}

In this work, we train two deep convolutional networks. One is
trained using tight face bounding boxes (DCNN$_{S}$), and the other
using large bounding boxes which include more contextual
(DCNN$_{L})$ information. In Section \ref{sec:exp}, we present
results which show that both networks capture discriminative
information and complement each other. In addition, the fusion of
two networks does significantly improve the final performance. The
architectures of both networks are
summarized in Tables~\ref{exp:deep_arch_S} and~\ref{exp:deep_arch_L}.\\

\indent Stacking small filters to approximate large filters and
building very deep convolutional networks reduces the number of
parameters but also increases the nonlinearity of the network as
discussed in \cite{simonyan_verydeep_2014,szegedy_going_2014}. In
addition, the resulting feature representation is compact and
discriminative. Therefore, for (DCNN$_{S}$), we use the same network
architecture presented in \cite{chen_unconstrained_2015} and train
it using the CASIA-WebFace dataset \cite{yi_learning_2014}. The
dimensionality of the input layer is $100 \times 100 \times 3$ for
RGB images. The network includes ten convolutional layers, five
pooling layers, and one fully connected layer. Each convolutional
layer is followed by a parametric rectified linear unit (PReLU)
\cite{he_delving_2015}, except the last one, conv52. Moreover, two
local normalization layers are added after conv12 and conv22,
respectively, to mitigate the effect of illumination variations. The
kernel size of all filters is $3 \times 3$. The first four pooling
layers use the max operator, and pool$_5$ uses average pooling. The
feature dimensionality of pool$_5$ is thus equal to the number of
channels of conv52 which is 320. The dropout ratio is set as 0.4 to
regularize Fc6 due to the large number of parameters (\emph{i.e.}
320 $\times$ 10548\footnote{The list of overlapping subjects is
available at
\url{http://www.umiacs.umd.edu/~pullpull/janus_overlap.xlsx}
\href{http://www.umiacs.umd.edu/~pullpull/janus_overlap.xlsx}{}} .).
The pool$_5$ feature is used for face representation. The extracted
features are further $L_2$-normalized to unit length before the
metric learning stage. If there are multiple images and frames
available for the subject template, we use the average of pool$_5$ features as the overall feature representation.\\

\begin{figure}[htp!]
\begin{center}
 \includegraphics[width=3in]{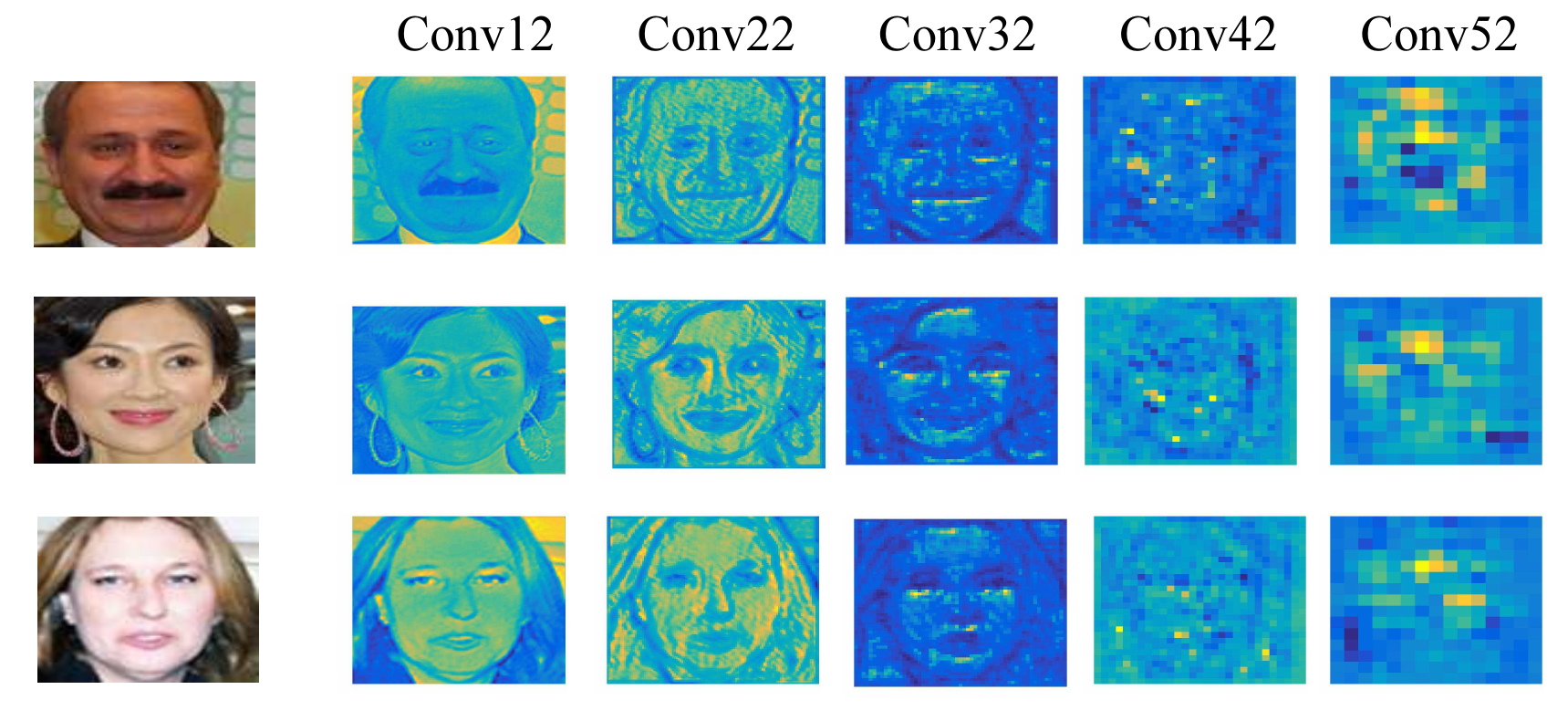}
\end{center}
  \caption{An illustration of some feature maps of conv12, conv22, conv32, conv42, and conv52 layers of DCNN$_{S}$ trained for the face identification task. At upper layers, the feature maps capture more global shape
features which are also more robust to illumination changes than
conv12. The feature maps are rescaled to the same size for
visualization purpose. The green pixels represent high activation
values, and the blue pixels represent low activation values as
compared to the green.}
  \label{fig:dcnn_activations}
\end{figure}

\begin{table*}[htb]
  \begin{minipage}[b]{0.5\textwidth}

    \centering \small
    \begin{tabular}{|c|c|c|c|c|c|c|c|}
    \hline
    Name & Type & Filter Size/Stride & $\#$Params\\
    \hline\hline
    conv11  & convolution & 3$\times$3 / 1 & 0.84K\\
    conv12  & convolution & 3$\times$3 / 1 & 18K\\
    pool1   & max pooling & 2$\times$2 / 2 &   \\
    conv21  & convolution & 3$\times$3 / 1 & 36K\\
    conv22  & convolution & 3$\times$3 / 1 & 72K\\
    pool2   & max pooling & 2$\times$2 / 2 &  \\
    conv31  & convolution & 3$\times$3 / 1 & 108K\\
    conv32  & convolution & 3$\times$3 / 1 &  162K\\
    pool3   & max pooling & 2$\times$2 / 2 &  \\
    conv41  & convolution & 3$\times$3 / 1 & 216K\\
    conv42  & convolution & 3$\times$3 / 1 & 288K\\
    pool4   & max pooling & 2$\times$2 / 2 &    \\
    conv51  & convolution & 3$\times$3 / 1 & 360K\\
    conv52  & convolution & 3$\times$3 / 1 & 450K\\
    pool5   & avg pooling & 7$\times$7 / 1 &   \\
    dropout & dropout (40\%) &             &  \\
    fc6     & fully connected &  10548    &  3296K\\
    loss    & softmax          &  10548    &   \\
    \hline\hline
    total   &                  &           &  5M \\
    \hline
    \end{tabular}
    \tabcaption{The architectures of DCNN$_S$.} \label{exp:deep_arch_S}
  \end{minipage}%
  \begin{minipage}[b]{0.5\textwidth}
    \centering \small
    \begin{tabular}{|c|c|c|c|}
    \hline
    Name & Type & Filter Size/Stride & $\#$Params\\
    \hline\hline
    conv1   & convolution      & 11$\times$11 / 4  & 35K\\
    pool1   & max pooling      & 3$\times$3 / 2 &     \\
    conv2   & convolution      & 5$\times$5 / 2 & 614K\\
    pool2   & max pooling      & 3$\times$3 / 2 &     \\
    conv3   & convolution      & 3$\times$3 / 2 & 885K\\
    conv4   & convolution      & 3$\times$3 / 2 & 1.3M\\
    conv5   & convolution      & 3$\times$3 / 1 &  885K\\
    conv6   & convolution      & 3$\times$3 / 1 &  590K\\
    pool6   & max pooling      & 3$\times$3 / 2 &     \\
    fc6     & fully connected & 1024 & 9.4M\\
        dropout & dropout (50\%) &             &  \\
    fc7     & fully connected & 512 & 524K\\
        dropout & dropout (50\%) &             &  \\
    fc8     & fully connected & 10548 & 5.5M\\
    loss    & softmax          &  10548    &   \\
            &  &  & \\
            &  &  & \\
            &  &  & \\
    \hline\hline
    total   &                  &           &  19.8M \\
    \hline
    \end{tabular}
    \tabcaption{The architecture of DCNN$_L$.} \label{exp:deep_arch_L}
  \end{minipage}
\end{table*}

% \subsection{Deep Convolutional Face Representation - Swami}
On the other hand, for DCNN$_{L}$, the deep network architecture
closely follows the architecture of the AlexNet
\cite{krizhevsky_imagenet_2012} with some notable differences:
reduced number of parameters in the fully connected layers; use of
Parametric Rectifier Linear units (PReLU's) instead of ReLU, since
they allow a negative value for the output based on a learnt
threshold and have been shown to improve the convergence rate
\cite{he_delving_2015}.

The reason for using the AlexNet architecure in the convolutional
layers is due to the fact that we initialize the convolutional layer
weights with weights from the AlexNet model which was trained using
the ImageNet challenge dataset. Several recent works
(\cite{transfer1},\cite{transfer2}) have empirically shown that this
transfer of knowledge across different networks, albeit for a
different objective, improves performance and more significantly
reduces the need to train using a large number of iterations. To
learn more domain specific information, we add an additional
convolutional layer, \textit{conv6} and initialize the fully
connected layers \textit{fc6-fc8} from scratch. Since the network is
used as a feature extractor, the last layer \textit{fc8} is removed
during deployment, thus reducing the number of parameters to 15M.
When the network is deployed. the features are extracted from
\textit{fc7} layers resulting in a dimensionality of 512. The
network is trained using the CASIA-WebFace dataset
\cite{yi_learning_2014}. The dimensionality of the input layer is
$227 \times 227 \times 3$ for RGB images.

In Figure \ref{fig:dcnn_activations}, we show some feature
activation maps of the DCNN$_{S}$ model. At upper layers, the
feature maps capture more global shape features which are also more
robust to illumination changes than Conv12 where the green pixels
represent high activation values, and the blue pixels represent low
activation values compared to the green.

\subsection{Triplet Similarity Embedding} \label{method:metric}

To further improve the performance of our deep features, we obtain a
low-dimensional discriminative projection of the deep features,
called the Triplet Similarity Embedding (TSE) that is learnt using
the training data provided for each split of IJB-A. The output of
the procedure is an embedding matrix $\mathbf{W} \in \mathbf{R}^{n
\times M}$ where $M$ is the dimensionality of the deep descriptor
(320 for DCNN$_S$ and 512 for DCNN$_L$) and we set $n=128$, thus
achieving dimensionality reduction in addition to an improvement in
performance.

In addition, for the TSE approach, the objective was two-fold (1) to
achieve as small dimensionality as possible for both networks (2) to
obtain a more discriminative representation in the low dimensional
space which means to push similar pairs together and dissimilar
pairs apart in the low-dimensional space. For learning $\mathbf{W}$,
we solve an optimization problem based on constraints involving
triplets - each containing two similar samples and one dissimilar
sample. Consider a triplet $\{a,p,n\}$, where $a$ (anchor) and $p$
(positive) are from the same class, but $n$ (negative) belongs to a
different class. Our objective is to learn a linear projection
$\mathbf{W}$ from the data such that the following constraint is
satisfied:
\begin{align}
(\mathbf{W}a)^{T} \cdot (\mathbf{W}p) > (\mathbf{W}a)^{T} \cdot
(\mathbf{W}n) \label{eq:constraint}
\end{align}

In our case, $\{a,p,n\} \in \mathbf{R}^{M}$ are deep descriptors
which are normalized to unit length. As such, $(\mathbf{W}a)^T \cdot
(\mathbf{W}p)$ is the dot-product or the similarity between ${a,p}$
under the projection $\mathbf{W}$. The constraint in
(\ref{eq:constraint}) requires that the similarity between the
anchor and positive samples should be higher than the similarity
between the anchor and negative samples in the low dimensional space
represented by $\mathbf{W}$. Thus, the mapping matrix $\mathbf{W}$
pushes similar pairs closer and dissimilar pairs apart, with respect
to the anchor point. By choosing the dimensionality of $\mathbf{W}$
as $n \times M$ where $n < M$, we achieve dimensionality reduction
in addition to better performance. For our work, we fix $n=128$
based on cross validation.

Given a set of labeled data points, we solve the following
optimization problem:
\begin{align}
\underset{\mathbf{W}}{\text{argmin}} \sum_{ {a,p,n} \in \mathbb{T}}
max(0,\alpha + a^T\mathbf{W^TW}n-a^T\mathbf{W^TW}p)
\label{eq:train1}
\end{align}
where $\mathbb{T}$ is the set of triplets and $\alpha$ is a margin
parameter chosen based on the validation set. In practice, the above
problem is solved in a Large-Margin framework using Stochastic
Gradient Descent (SGD) and the triplets are sampled online. The
update step for solving (\ref{eq:train1}) with SGD is:

\begin{align}
\mathbf{W}_{t+1} = \mathbf{W}_t - \eta * \mathbf{W}_t * (a(n-p)^T  +
(n-p)a^T) \label{eq:update}
\end{align}
where $\mathbf{W}_t$ is the estimate at iteration $t$,
$\mathbf{W}_{t+1}$ is the updated estimate, $\{a,p,n\}$ is the
triplet sampled at the current iteration and $\eta$ is the learning
rate which is set to 0.01 for the current work.

The entire procedure takes 3-5 minutes per split using a standard
C++ implementation. More details regarding the optimization
algorithm can be found in \cite{tse}. At each iteration, we sample
1000 instances from the whole training set to choose the negatives.
Since the training set is relatively small for the datasets
considered in this experiment, the entire training set is held in
memory. Going forward this could be made efficient by using a buffer
which will be replenished periodically, thus requiring a constant
memory requirement. The computational complexity of each iteration
is $O(M^2)$, that is, the complexity varies quadratically with the
dimension of the deep descriptor. The technique closest to the one
presented in this section, which is used in recent works
(\cite{parkhi_deep_2015},\cite{schroff_facenet_2015}) computes the
embedding $\mathbf{W}$ based on satisfying the distance constraints
given below:
\begin{align}
\underset{\mathbf{W}}{\text{argmin}} \sum_{ {a,p,n} \in \mathbb{T}} max\{0,\alpha + (a-p)^T\mathbf{W^TW}(a-p)- \\
                (a-n)^T\mathbf{W^TW}(a-n) \}
\label{eq:dist}
\end{align}
To be consistent with the terminology used in this paper, we call it
Triplet Distance Embedding (\textbf{TDE}). It should be noted that
the \textbf{TSE} formulation is different from \textbf{TDE}, in
that, the current work uses inner-product based constraints between
triplets to optimize for the embedding matrix as opposed to
norm-based constraints used in the \textbf{TDE} method. To choose
the dimensionality, we test the values 64,128,256 using a 5 fold
validation scheme for each split. The learning rate is chosen as
0.02 and is fixed throughout the procedure. The margin parameter is
chosen as 0.1. We find from our experiments that the lower margin
works better but since we perform hard negative mining at each step,
the method is not particularly sensitive to the margin parameter.

In general, to learn a reasonable distance measure directly using
pairwise or triplet metric learning approach requires huge amount of
data (\emph{i.e.,}, the state-of-the-art
approach~\cite{schroff_facenet_2015} uses 200M images). In addition,
the proposed approach decouples the DCNN feature learning and metric
learning steps due to memory constraints. To learn a reasonable
distance measure requires generating the informative pairs or
triplets. The batch size used for SGD is limited by the memory size
of the graphics card. If the model is trained end-to-end, then only
a small batch size is available for use. Thus, in this work, we
perform DCNN model training and metric learning independently. In
addition, for the publicly available deep
model~\cite{parkhi_deep_2015}, it is also trained first with softmax
loss and followed by finetuning the model with verification loss
while freezing the convolutional and fully connected layers except
the last one so that the transformation which is equivalent to the
proposed approach can be learned.

\section{Experimental Results} \label{sec:exp}
In this section, we present the results of the proposed automatic
system for both face detection and face verification tasks on the
challenging IARPA Janus Benchmark A (IJB-A)
~\cite{klare_janus_2015}, its extended version Janus Challenging set
2 (JANUS CS2) dataset, and the LFW dataset. The JANUS CS2 dataset
contains not only the sampled frames and images in the IJB-A, but
also the original videos. In addition, the JANUS CS2
dataset\footnote{The JANUS CS2 dataset is not publicly available
yet.} includes considerably more test data for identification and
verification problems in the defined protocols than the IJB-A
dataset. The receiver operating characteristic curves (ROC) and the
cumulative match characteristic (CMC) scores are used to evaluate
the performance of different algorithms for face verification. The
ROC curve measures the performance in verification scenarios, where
the vertical axis is true acceptance rate (TAR) which represents the
degree to correctly match the face image (\emph{i.e.}, deep
features) from the same person and the horizontal axis shows false
acceptance rate (FAR) which represents the degree to falsely match
the biometric information from one person to another. The CMC score
measures the accuracy in closed set identification scenarios.

\subsection{Face Detection on IJB-A}

\begin{figure}[htp!]
\centering
\includegraphics[height=2.5in]{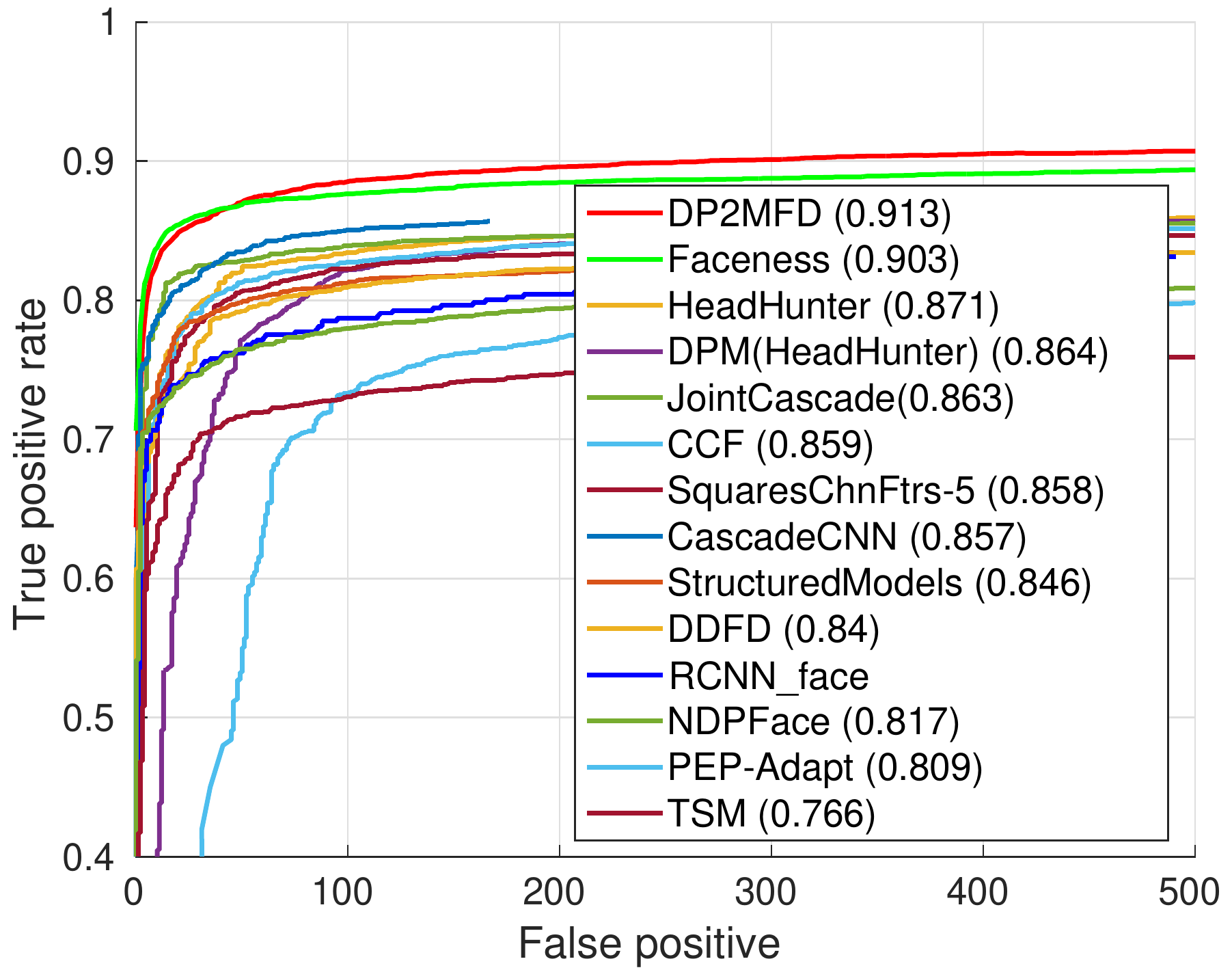}
\vspace{+1mm} \caption{Face detection performance evaluation on the
FDDB dataset~\cite{ranjan_deep_2015}.} \label{exp:fddb_detection}
\end{figure}

\begin{figure*}[htp!]
\centering \subfigure[]{
\includegraphics[height=2.5in]{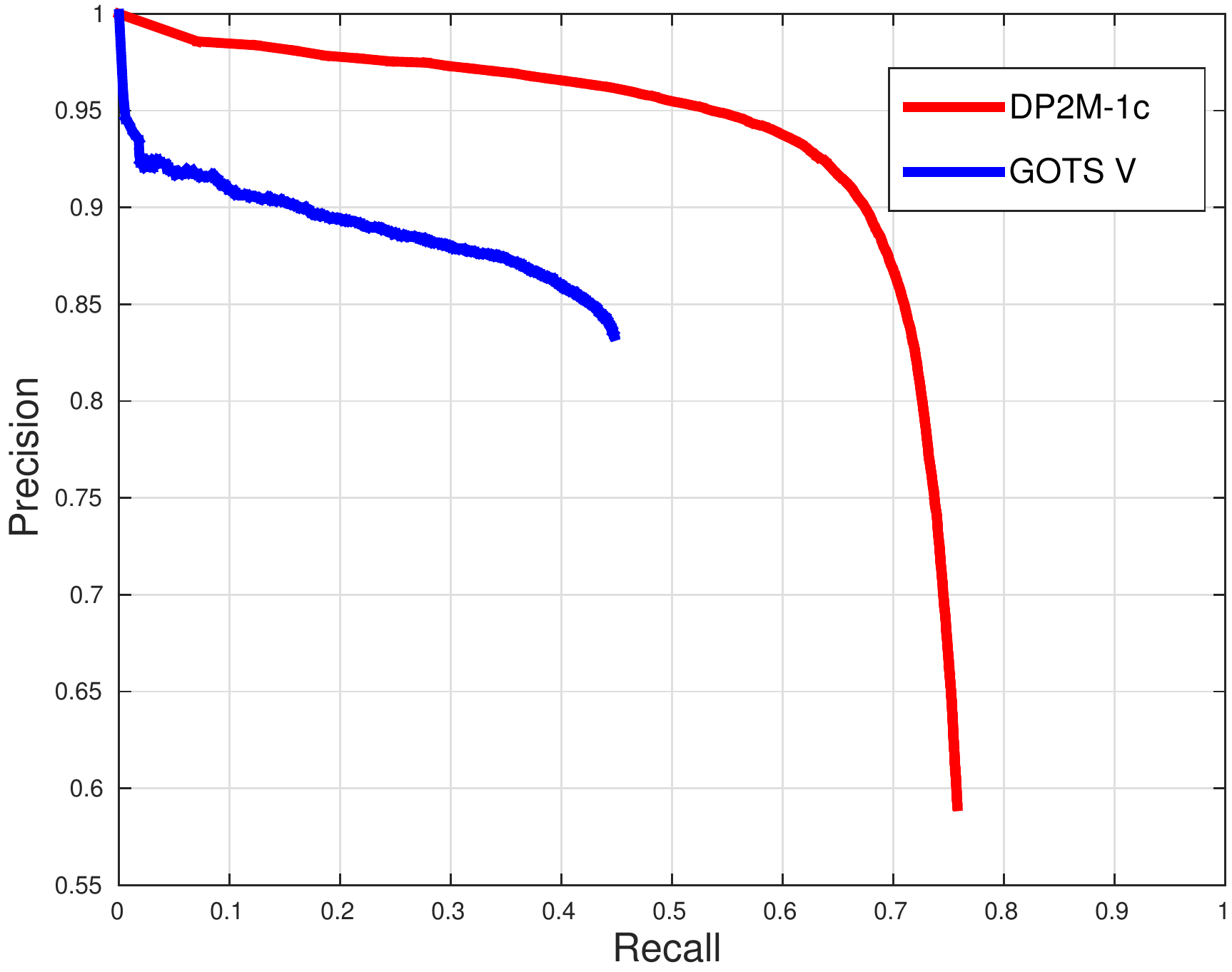}
} \subfigure[]{
\includegraphics[height=2.5in]{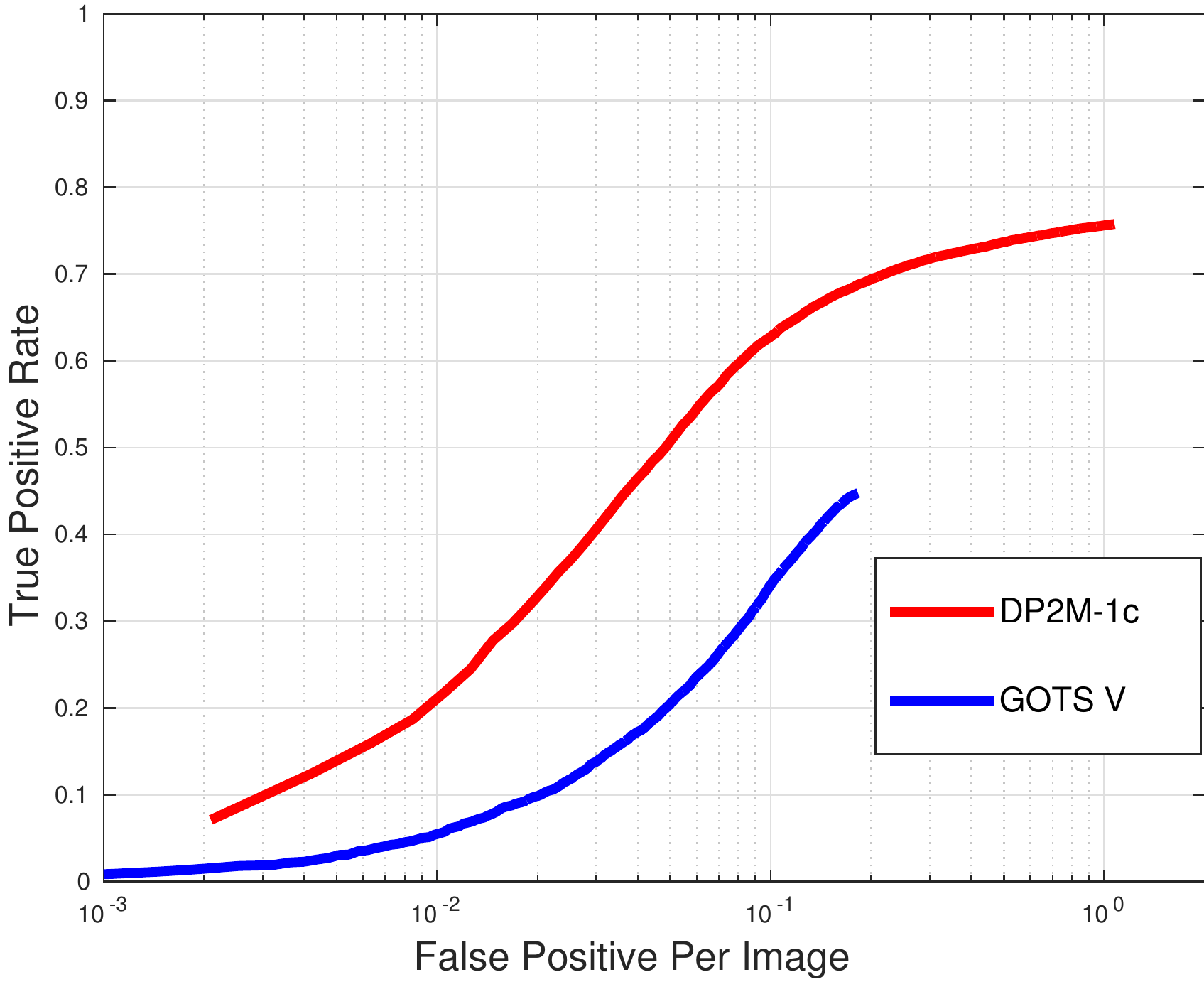}
} \vspace{+1mm} \caption{Face detection performance evaluation on
the IJB-A dataset: (a) Precision vs. recall curves. (b) ROC
curves~\cite{ranjan_deep_2015}.} \label{exp:ijba_face_pr_roc}
\end{figure*}

\begin{figure}[htp!]
\centering \subfigure[]{
\includegraphics[height=1.1in]{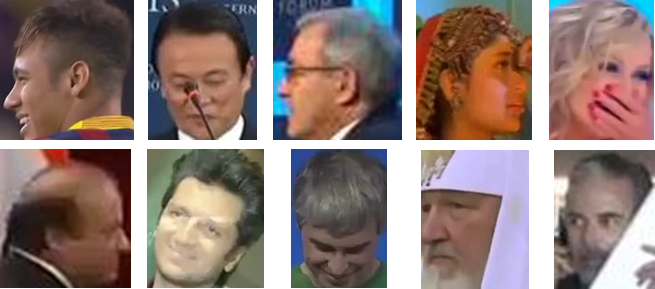}
} \subfigure[]{
\includegraphics[height=1.1in]{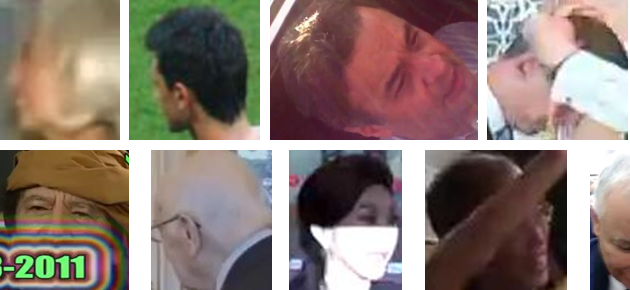}
} \vspace{+1mm} \caption{(a) shows the difficult faces in the IJB-A
dataset that are successfully detected by DP2MFD, and (b) shows
faces that are not detected by DP2MFD. From the results, we can see
that DP2MFD can handle difficult occlusion, partial face, large
illumination and pose variations.} \label{exp:misdetection}
\end{figure}

The IJB-A dataset contains images and sampled video frames from 500
subjects collected from online media \cite{klare_janus_2015},
\cite{cheney_unconstrained_2015}. For face detection task, there are
67,183 faces of which 13,741 are from images and the remaining are
from videos. The locations of all faces in the IJB-A dataset have
been manually annotated. The subjects were captured so that the
dataset contains wide geographic distribution. Nine different face
detection algorithms were evaluated on the IJB-A dataset
\cite{cheney_unconstrained_2015}, and the algorithms compared in
\cite{cheney_unconstrained_2015} include one commercial off the
shelf (COTS) algorithm, three government off the shelf (GOTS)
algorithms, two open source face detection algorithms (OpenCV's
Viola Jones and the detector provided in the Dlib library), and GOTS
ver 4 and 5. In Figure \ref{exp:ijba_face_pr_roc}, we show the
precision-recall (PR) curves and the ROC curves, respectively
corresponding to the method used in our work and one of the best
reported methods in \cite{cheney_unconstrained_2015}. We see that
the face detection algorithm used in our system outperforms the best
performing method reported in \cite{cheney_unconstrained_2015} by a
large margin. In Figure \ref{exp:misdetection} (b), we illustrate
typical faces in the IJB-A dataset that are not detected by DP2MFD,
and we can find the faces to be usually in very extreme conditions
which contain limited information for face verification. However, in
Figure \ref{exp:misdetection} (a), we also show that the DP2MFD
algorithm can handle very difficult faces but relatively reasonable
as compared to those in \ref{exp:misdetection} (b). As shown in
Figure~\ref{exp:fddb_detection}, the DP2MFD algorithm also achieves
top performance in the challenging FDDB benchmark \cite{fddbTech}
for face detection with a large performance margin compared to most
algorithms. Some of the recent published methods compared in the
FDDB evaluation include Faceness\cite{faceness_ICCV2015}, HeadHunter
\cite{HeadHunter_Mathias_ECCV2014}, JointCascade
\cite{JointCascade_LI_ECCV2014}, CCF \cite{CCF_ICCV2015}, Squares-
ChnFtrs-5 \cite{HeadHunter_Mathias_ECCV2014}, CascadeCNN
\cite{CascadeCNN_CVPR2015}, Structured Models \cite{Yan2014790},
DDFD \cite{DDFD_ICMR2015}, NDPFace \cite{NPDFace_PAMI2015},
PEP-Adapt \cite{PEP_Adapt_LI_ICCV2013} and TSM \cite{zhu2012face}.
More comparison results with other face detection data sets are
available in \cite{ranjan_deep_2015}. Since the CS2 dataset has not
been released to public, we are not able to provide comparisons with
other existing face detectors.


\subsection{Facial Landmark Detection on IJB-A}
\begin{figure}[htp!]
 \centering
\includegraphics[width=8.5cm]{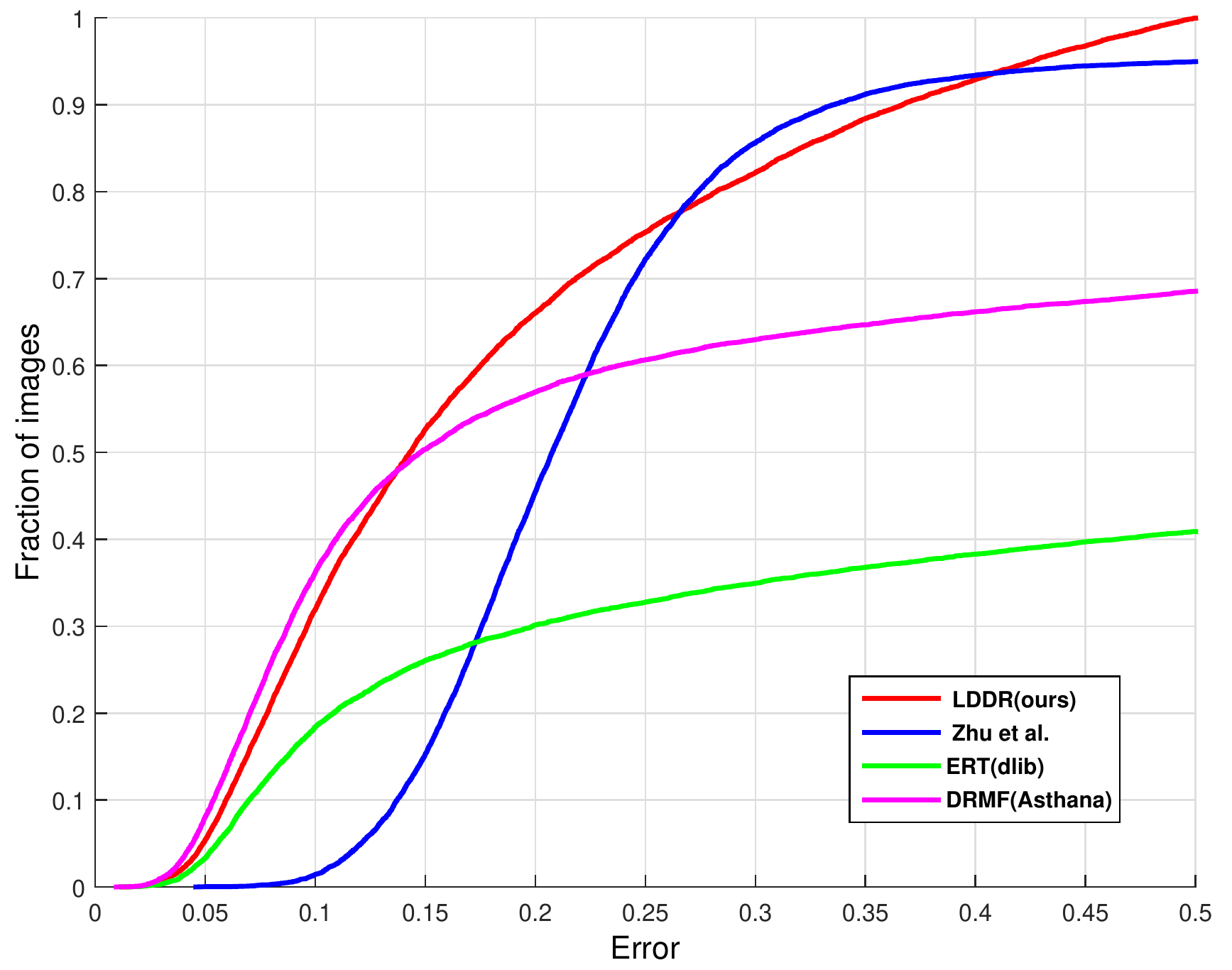}
\caption{Average 3-pt error (normalized by eye-nose distance) vs
fraction of images in the IJB-A dataset~\cite{kumar_face_2016}.}
\label{fig:Errors_IJBA}
\end{figure}

\begin{figure}[tb]
\begin{center}
 \includegraphics[width=2.3in]{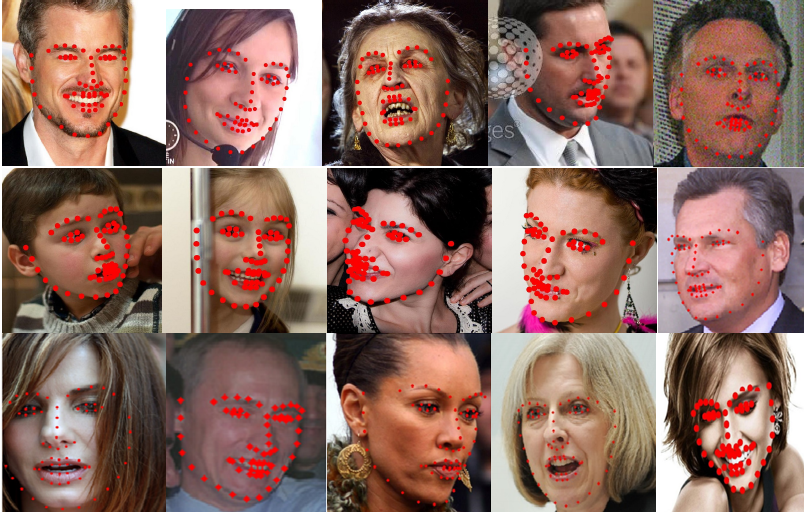}
\end{center}
  \caption{Sample facial landmark detection results.}
  \label{exp:face_landmark}
\end{figure}

We also evaluate the performance of our facial landmark detection
method on the IJB-A dataset. For the training data, we take 3148
images in total from the LFPW \cite{belhumeur2013localizing}, Helen
\cite{le2012interactive} and AFW \cite{zhu2012face} datasets and
test on the IJB-A dataset. The subjects were captured so that the
dataset contains wide geographic distribution. The challenge comes
through the wide diversity in pose, illumination and resolution. Our
method produces 68 facial landmark points following MultiPIE
\cite{gross2010multi} markup format. We evaluate the performance
using the Normalized Mean Square Error and average pt-pt error
(normalized by face size) vs fraction of images plots of different
methods. Since IJB-A is annotated only with 3 key-points on the
faces (two eyes and nose base) by human annotators, the interoccular
distance error was normalized by the distance between nose tip and
the midpoint of the eye centers. In Figure \ref{fig:Errors_IJBA}, we
present a comparison of our algorithm with \cite{zhu2012face},
\cite{asthana2013robust} and \cite{kazemi2014one}. For the Helen
dataset, we show the performance of 49-point and full 68-point
results in Table~\ref{tbl:Helen}. Our deep descriptor-based global
shape regression method outperforms the above mentioned
state-of-the-art methods in both high-quality (Helen) and
low-quality (IJB-A) images. Samples of detected landmarks results
are shown in Figure \ref{exp:face_landmark}. More evaluation results
for landmark detection on other standard datasets may be found in
\cite{kumar_face_2016}. \indent Once facial landmark detection is
completed, we choose seven landmark points (\emph{i.e.} two left eye
corners, two right eye corners, nose tip, and two mouth corners) out
of the detected 68 points and apply the similarity transform to warp
the faces into canonical coordinates.

\begin{table}[thp!]
\begin{center}
\resizebox{\linewidth}{!}{%
\begin{tabular}{|p{4.1cm}|p{1.5cm}|p{1.5cm}|}
\hline
\centering {\it Method} & \centering {\it 68-pts} &  {\it 49-pts}\\
\hline\hline
\centering Zhu \it{et al.} \cite{zhu2012face}   & \centering 8.16         & 7.43  \\
\centering DRMF \cite{asthana2013robust}         &\centering 6.70         & -      \\
\centering RCPR \cite{RCPR}         &\centering 5.93         & 4.64  \\
\centering SDM \cite{6618919}          & \centering 5.50         & 4.25       \\
\centering GN-DPM \cite{6909635}       & \centering 5.69         & 4.06    \\
\centering CFAN \cite{DBLP:conf/eccv/ZhangSKC14}       & \centering 5.53         & -        \\
\centering CFSS \cite{Zhu_2015_CVPR}       & \centering {\bf 4.63}         & 3.47        \\
\hline
\centering {\bf LDDR(Ours)}   & \centering  4.76   & {\bf 2.36}\\
\hline
\end{tabular}}
\vskip 4pt \caption{Averaged error comparison of different methods
on the Helen dataset~\cite{kumar_face_2016}.} \label{tbl:Helen}
\end{center}
\end{table}

\subsection{IJB-A and JANUS CS2 for Face Verification}
\begin{figure*}[htp!]
\begin{center}
 \includegraphics[width=4in]{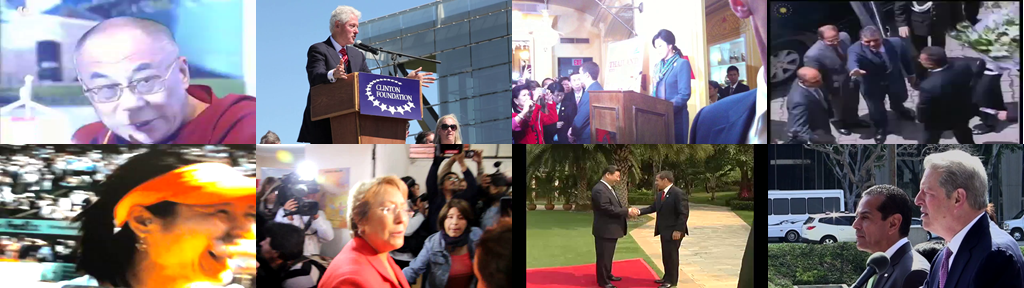}
\end{center}
  \caption{Sample images and frames from the IJB-A (top) and JANUS CS2 datasets (bottom). Challenging variations
  due to pose, illumination, resolution, occlusion, and image quality are present in these images. }
  \label{exp:sample_janus}
\end{figure*}

For face verification task, both IJB-A and JANUS CS2 datasets
contain 500 subjects with 5,397 images and 2,042 videos split into
20,412 frames, 11.4 images and 4.2 videos per subject. Sample images
and video frames from the datasets are shown in
Figure~\ref{exp:sample_janus}. (\emph{i.e.}, the videos are only
released for the JANUS CS2 dataset.) The IJB-A evaluation protocol
consists of verification (1:1 matching) over 10 splits. Each split
contains around 11,748 pairs of templates (1,756 positive and 9,992
negative pairs) on average. Similarly, the identification (1:N
search) protocol also consists of 10 splits, which are used to
evaluate the search performance. In each search split, there are
about 112 gallery templates and 1,763 probe templates (\emph{i.e.}
1,187 genuine probe templates and 576 impostor probe templates). On
the other hand, for the JANUS CS2, there are about 167 gallery
templates and 1,763 probe templates and all of them are used for
both identification and verification. The training set for both
datasets contains 333 subjects, and the test set contains 167
subjects without any overlapping subjects. Ten random splits of
training and testing are provided by each benchmark, respectively.
The main differences between IJB-A and JANUS CS2 evaluation
protocols are that (1) IJB-A considers the open-set identification
problem and the JANUS CS2 considers the closed-set identification
and (2) IJB-A considers the more difficult pairs which are subsets
of JANUS CS2 dataset.

Unlike the LFW and YTF datasets, which only use a sparse set of
negative pairs to evaluate the verification performance, the IJB-A
and JANUS CS2 datasets divide the images/video frames into gallery
and probe sets so that all the available positive and negative pairs
are used for the evaluation. Also, each gallery and probe set
consist of multiple templates. Each template contains a combination
of images or frames sampled from multiple image sets or videos of a
subject. For example, the size of the similarity matrix for JANUS
CS2 split1 is 167 $\times$ 1806 where 167 are for the gallery set
and 1806 for the probe set (\emph{i.e.} the same subject reappears
multiple times in different probe templates). Moreover, some
templates contain only one profile face with a challenging pose with
low quality imagery. In contrast to LFW and YTF datasets, which only
include faces detected by the Viola Jones face detector
\cite{viola_robust_2004}, the images in the IJB-A and JANUS CS2
contain extreme pose, illumination, and expression variations. These
factors essentially make the IJB-A and JANUS CS2 challenging face
recognition datasets \cite{klare_janus_2015}.

\begin{figure*}[htp!]
\centering \subfigure[]{
\includegraphics[height=2.5in]{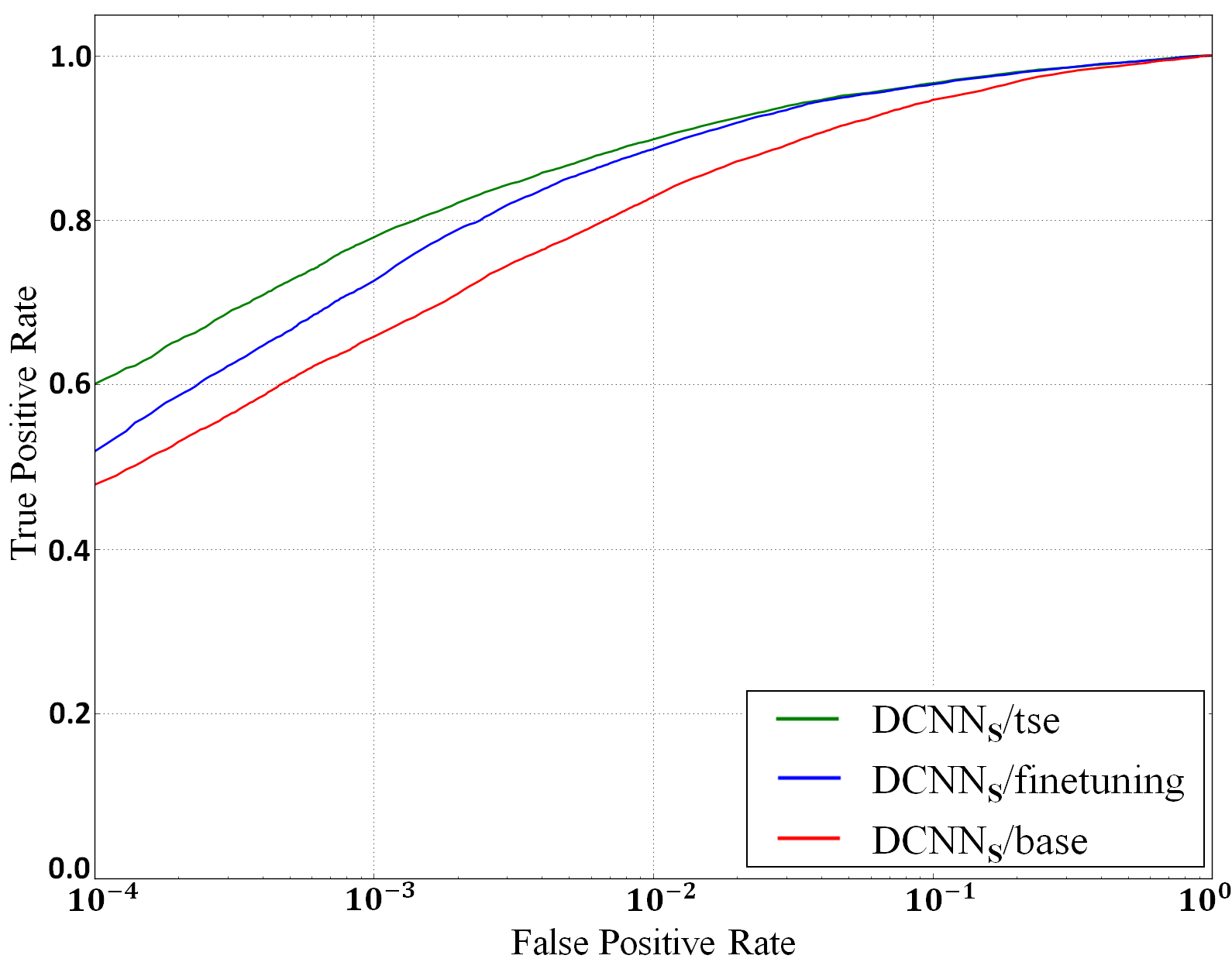}
} \subfigure[]{
\includegraphics[height=2.5in]{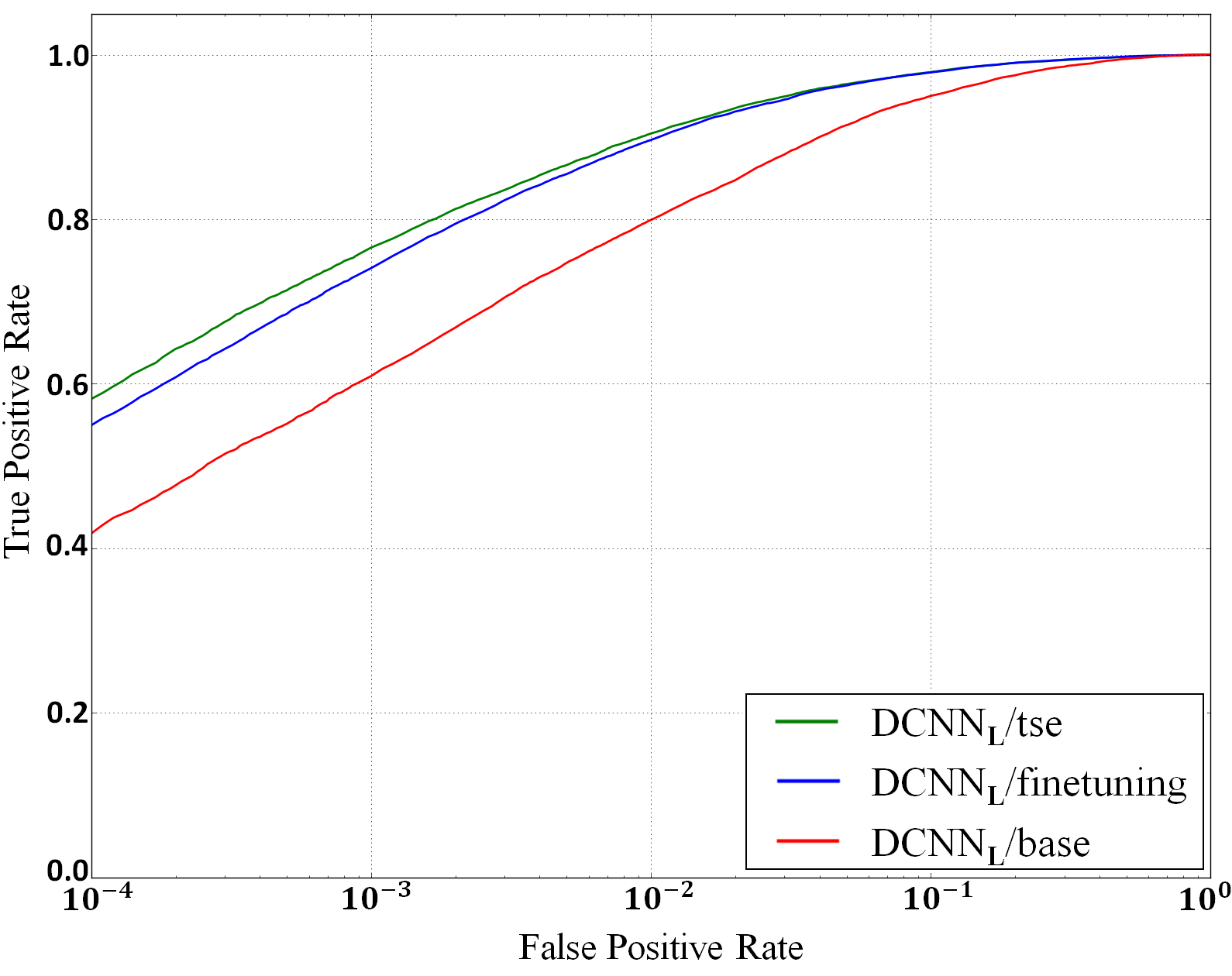}
} \vspace{+1mm} \caption{The performance evaluation for face
verification tasks of (a) DCNN$_S$ and (b) DCNN$_L$ of before
finetuning, with finetuning, and with finetuning and triplet
similarity embedding for the JANUS CS2 dataset under Setup 3
(semi-automatic mode). Fine tuning is done only using the training
data in each split.} \label{exp:cs2_jc_swami}
\end{figure*}

\begin{figure*}[htp!]
\centering \subfigure[]{
\includegraphics[height=2.5in]{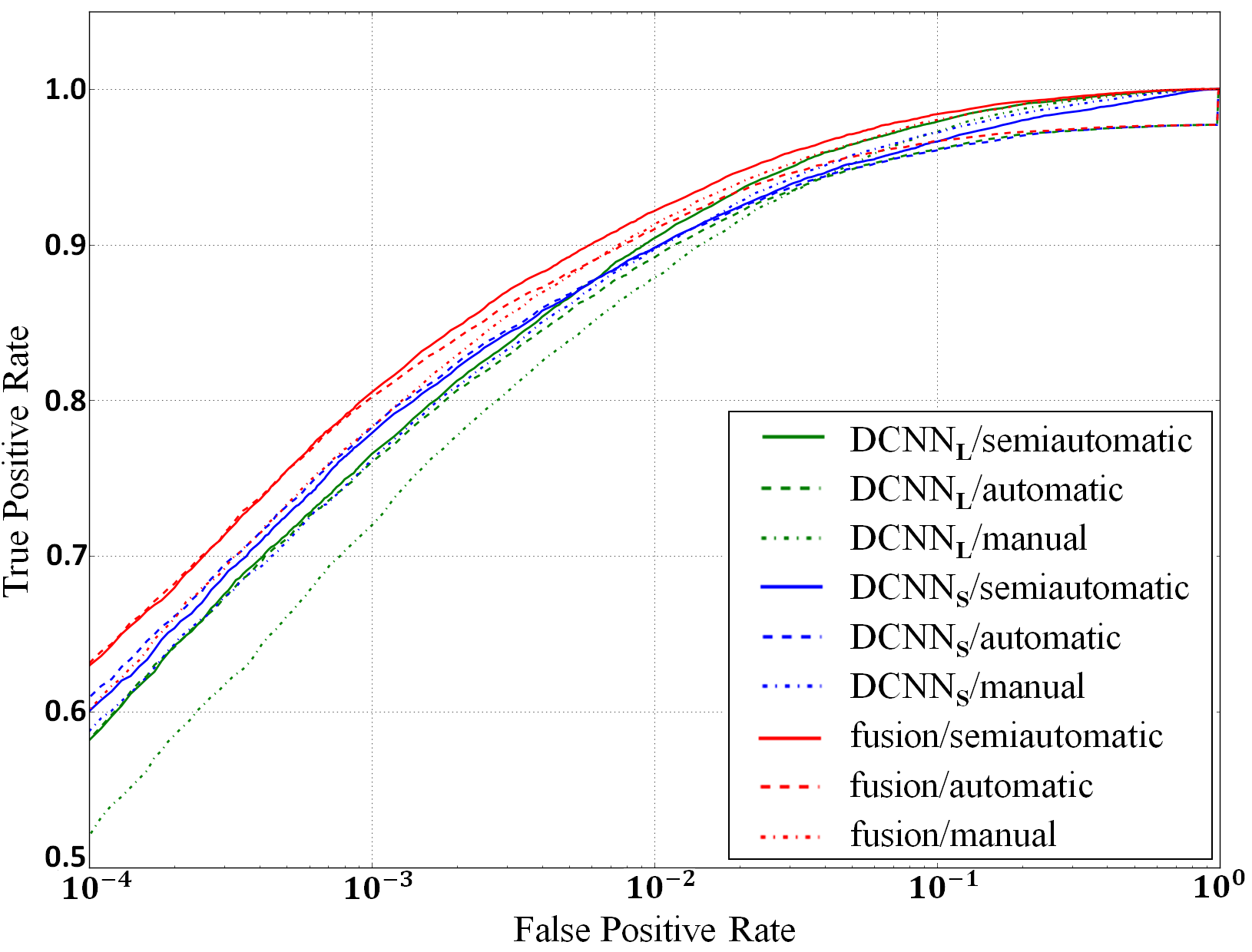}
} \subfigure[]{
\includegraphics[height=2.5in]{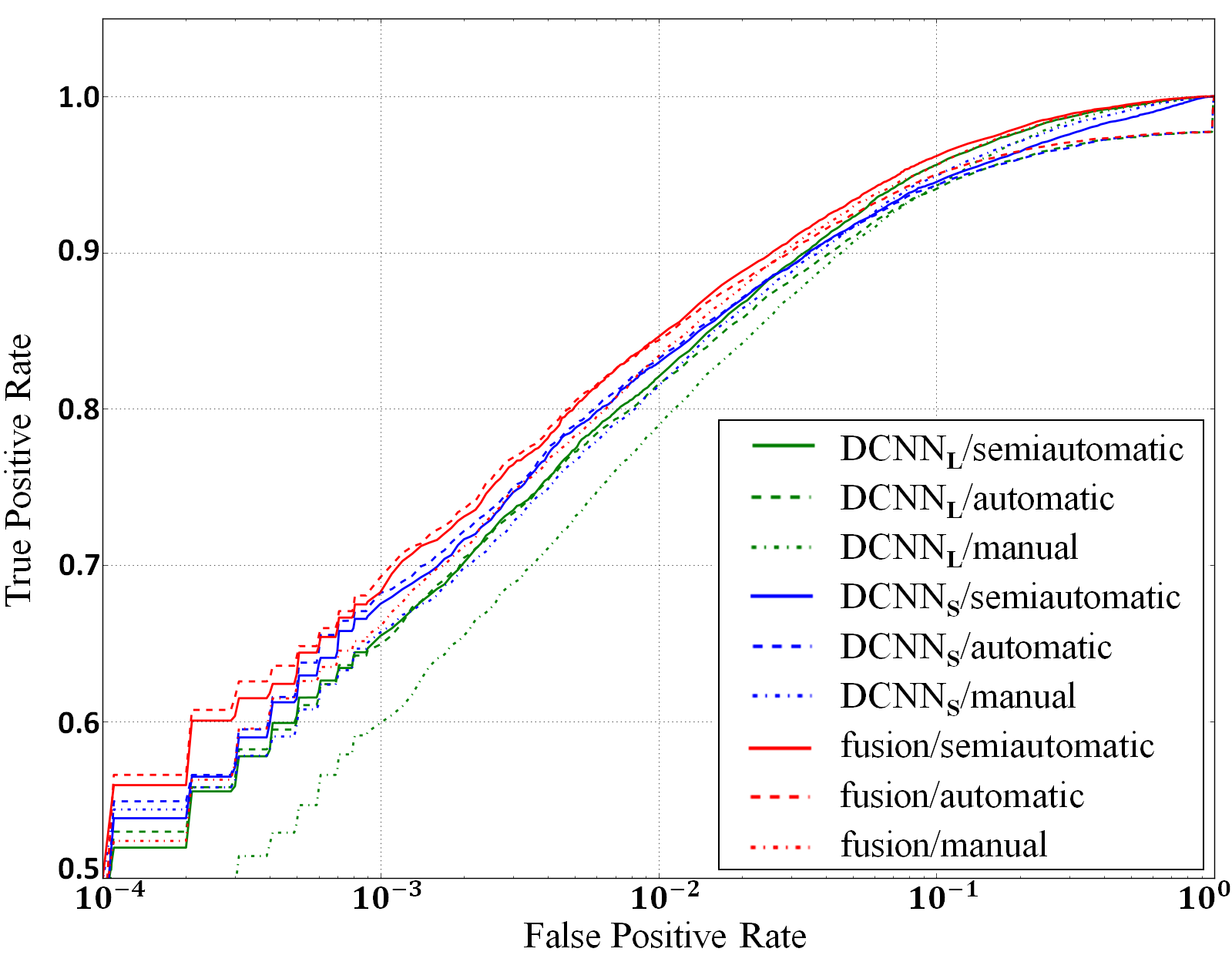}
} \vspace{+1mm}\\ \subfigure[]{
\includegraphics[height=2.5in]{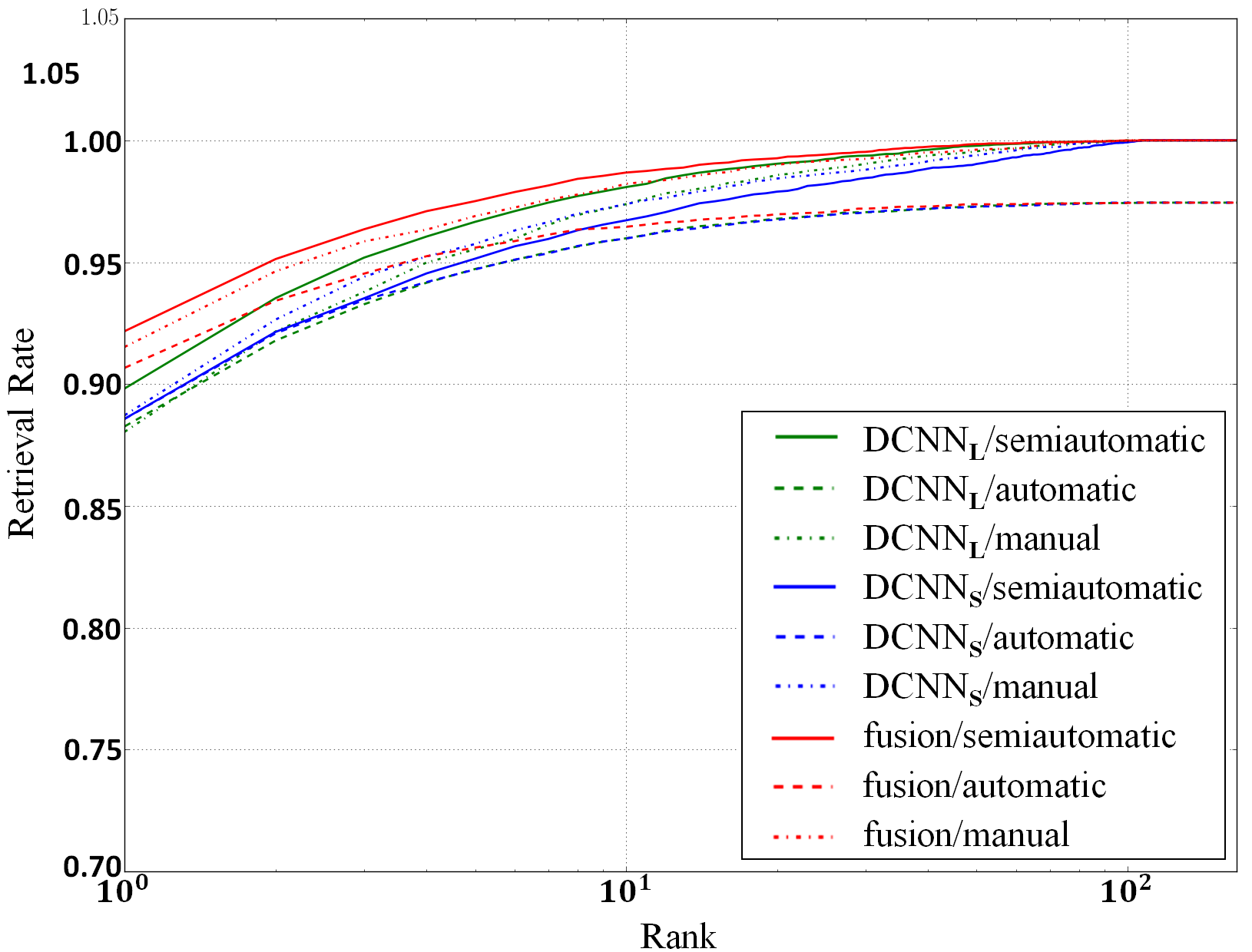}
} \vspace{+1mm} \caption{(a) and (b) show the face verification
performance of the fusion model for JANUS CS2 and IJB-A (1:1)
verification, respectively, and (c) shows the face identification
performance of the fusion model for IJB-A (1:N) identification for
all the three setups. Fine tuning is done only using the training
data in each split.} \label{exp:cs2_jc_swami_ijba}
\end{figure*}
\subsection{Performance Evaluations of Face Verification on IJB-A and JANUS CS2}
%\subsection{Implementation Details for the End-to-End Automated System - from ITA paper}

%setup 1: manual. Manual 3 landmarks, if 3 landmarks simultaneously
%available use the face rectangle given by the meta data.
%
%setup 2: is auto directory. In this setup when we get a video we use
%the face association method and to extract the bouding box to
%perform detection and fiducial using the face association. If it is
%an image we perform detection and keypoint detection independently.
%If for every image or frame in a template we are unable to detect
%the target person, then we concede, that is we assign it the lowest
%similarity score possible.
%
%setup 3:  is semi-automatic. In this setup if we are able to detect
%the target person in an image then we follow setup 2, if not we
%follow setup 1.
%
%Plots: manual, automatic and semi automatic:  CS2 ver, IJB-A ver,
%IJB-A search cmc, FTTL, error bars
%
%improvement CS2 verify 2 curves: base, ft, fttl for each method
%
%IJB-A verification: semi-auto fused, vs all performers
%
%Tables: IJB-A search, FTTL fusion CMC numbers, 1,5, 10, 20
To take different situations into account, we have considered three
modes of evaluations, manual, automatic and semi-automatic modes.
This enables the handling of cases where we are unable to detect any
of the faces (\emph{i.e.,} the failure of face detection.) in the
images of the given template and also to compare the performance
with the one using the metadata provided with the dataset. We
describe the setups of performance evaluation in details as follows:
\begin{itemize}
  \item \textbf{Setup 1 (manual mode):} Under this setup, we directly use
    the three facial landmarks and face bounding boxes provided along with the
    datasets.
  \item \textbf{Setup 2 (automatic mode):} In this setup when we get a video we use
    the face association method to detect and track the faces and to
    extract the bounding box to perform fiducial detection. If it is an
    image, we perform detection and facial landmark detection
    independently. For every image or frame in a template in which we are unable to
    detect the target face, we are unable to compare the template with others and thus assign all the corresponding entries for the template in the similarity
    matrices to the lowest similarity scores, -Inf.
  \item \textbf{Setup 3 (semi-automatic mode):} In this setup if we are able to detect
    the target face in an image then we follow setup 2. Otherwise, we follow setup 1 to use
    the metadata of the dataset for the faces which are not detected and tracked by our algorithms.
\end{itemize}

To evaluate the performance of two networks individually, we present
the ROC curves of DCNN$_S$ and DCNN$_L$ of the Setup 3 (\emph{i.e.},
semi-automatic mode) for the JANUS CS2 dataset in
Figure~\ref{exp:cs2_jc_swami}. As shown in the figures, the
performances are consistently improved for both networks after
fine-tuning the models previously trained using the CASIA-WebFace
dataset on the training data of JANUS CS2. Triplet similarity
embedding (TSE) further increase the performance for both networks,
especially for the TAR number at the low FAR interval. For all the
results presented here, fine tuning is done using only the training
data in each split. The gallery dataset is not used for parameter
fine tuning or for triplet similarity embedding. Then, we perform
the fusion of the two networks by adding the corresponding
similarity scores together and demonstrate the fusion results of all
the three setup for the verification task of both JANUS CS2 and
IJB-A in Figure ~\ref{exp:cs2_jc_swami_ijba} (a) and (b),
respectively. In Figure ~\ref{exp:cs2_jc_swami_ijba} (c), we present
the CMC curve for the IJB-A identification task. From
Figure~\ref{exp:cs2_jc_swami_ijba},
 it can be seen that even the simple fusion strategy used in this work significantly boosts the performance. Since DCNN$_S$ is
trained using tight face bounding boxes (DCNN$_{S}$) and DCNN$_L$
using the large ones which includes more context (DCNN$_{L})$, one
possible reason for the performance improvement is that the two
networks contain discriminative information learned from different
scales and complement each other. In addition, the figure also shows
that the performance of our system in Setup 2 (the automatic mode)
is comparable to Setup 1 (the manual mode) and Setup 3 (the
semi-automatic mode). This demonstrates the robustness of each
component of our system.

\begin{table*}[htp!]
\centering \scriptsize
\begin{tabular}{|c|c|c|c|c|c|c|}
  \hline
  IJB-A-Verif & DCNN (setup 1) & DCNN (setup 2) & DCNN (setup 3) & DCNN$_{m}$ (setup 1)& DCNN$_{m}$ (setup 2)& DCNN$_{m}$ (setup 3)\\
  \hline
  FAR=1e-2 & 0.834 $\pm$ 0.036 & 0.844 $\pm$ 0.026 & 0.846 $\pm$ 0.029 & 0.863 $\pm$ 0.02& 0.885 $\pm$ 0.014 & \textbf{0.889 $\pm$ 0.016}\\
  FAR=1e-1 & 0.956 $\pm$ 0.008 & 0.95  $\pm$ 0.005 & 0.962 $\pm$ 0.007 & 0.966 $\pm$ 0.05& 0.954 $\pm$ 0.003 & \textbf{0.968 $\pm$ 0.005}\\
  \hline\hline
  IJB-A-Ident & DCNN (setup 1) & DCNN (setup 2) & DCNN (setup 3) & DCNN$_{m}$ (setup 1)& DCNN$_{m}$ (setup 2)& DCNN$_{m}$ (setup 3)\\
  \hline
  Rank-1  & 0.915 $\pm$ 0.011  & 0.907 $\pm$ 0.011 & 0.922 $\pm$ 0.011& 0.916 $\pm$ 0.009 & 0.923 $\pm$ 0.01 &\textbf{0.942 $\pm$ 0.008}\\
  Rank-5  & 0.969 $\pm$ 0.007 & 0.955 $\pm$ 0.007  & 0.975 $\pm$ 0.006 & 0.971 $\pm$ 0.007 & 0.961 $\pm$ 0.006&\textbf{0.98  $\pm$ 0.005}\\
  Rank-10 & 0.982 $\pm$ 0.005 & 0.965 $\pm$ 0.005  & 0.987 $\pm$ 0.001 & 0.981 $\pm$ 0.005&  0.969 $\pm$ 0.004&\textbf{0.988 $\pm$ 0.003}\\
  \hline
  \hline
  IJB-A-Ident & DCNN (setup 1) & DCNN (setup 2) & DCNN (setup 3) & DCNN$_{m}$ (setup 1)& DCNN$_{m}$ (setup 2)& DCNN$_{m}$ (setup 3)\\
  \hline
  FPIR=0.01 & 0.618 $\pm$ 0.05 & 0.64 $\pm$ 0.043  & 0.631 $\pm$ 0.041 & 0.639 $\pm$ 0.057 & 0.646 $\pm$ 0.055 &\textbf{0.654 $\pm$ 0.001}\\
  FPIR=0.1  & 0.799 $\pm$ 0.014 & 0.806 $\pm$ 0.012  & 0.813 $\pm$ 0.014 & 0.816 $\pm$ 0.015 & 0.827 $\pm$ 0.012 &\textbf{0.836 $\pm$ 0.01}\\
  \hline
\end{tabular}
 \vspace{+1mm} \caption{Results on the IJB-A
dataset. The TAR of all the approaches at FAR=0.1 and 0.01 for the
ROC curves (IJB-A 1:1 verification). The Rank-1, Rank-5, and Rank-10
retrieval accuracies of the CMC curves and TPIR at FPIR = 0.01 and
0.1 (IJB-A 1:N identfication). We also show the results before and
after media averaging where $m$ means media averaging.}
\label{exp:roc_cmc_scores_ijba_media}
\end{table*}

\begin{table*}[htp!]
\centering \scriptsize
\begin{tabular}{|c|c|c|c|c|c|c|}
  \hline
  CS2-Verif&DCNN (setup 1)&DCNN (setup 2) & DCNN (setup 3)& DCNN$_{m}$ (setup 1)& DCNN$_{m}$ (setup 2)& DCNN$_{m}$ (setup 3)\\
  \hline
  FAR=1e-2& 0.913 $\pm$ 0.008 & 0.91 $\pm$  0.008  &  0.922 $\pm$ 0.007 & 0.92  $\pm$ 0.01  & 0.922 $\pm$ 0.008& \textbf{0.935 $\pm$ 0.007}\\
  FAR=1e-1& 0.98  $\pm$ 0.004 & 0.967 $\pm$ 0.003  &  0.984 $\pm$ 0.003 & 0.981 $\pm$ 0.003 & 0.968 $\pm$ 0.003& \textbf{0.986 $\pm$ 0.002}\\
  \hline
  CS2-Ident&DCNN (setup 1)&DCNN (setup 2)& DCNN (setup 3)& DCNN$_{m}$ (setup 1)& DCNN$_{m}$ (setup 2)& DCNN$_{m}$ (setup 3)\\
  \hline
  Rank-1& 0.9 $\pm$ 0.01&0.896 $\pm$ 0.008 & 0.909 $\pm$ 0.008 & 0.905 $\pm$ 0.007   & 0.915 $\pm$ 0.007& \textbf{0.931 $\pm$ 0.007}\\
  Rank-5& 0.963 $\pm$ 0.006&0.954 $\pm$ 0.006 & 0.969 $\pm$ 0.006 & 0.965 $\pm$ 0.004& 0.959 $\pm$ 0.005& \textbf{0.976 $\pm$ 0.004}\\
  Rank-10&0.977 $\pm$ 0.006&0.965 $\pm$ 0.004 & 0.981 $\pm$ 0.003& 0.977 $\pm$ 0.004 & 0.967 $\pm$ 0.004& \textbf{0.985 $\pm$ 0.002}\\
  \hline
\end{tabular}
\vspace{+1mm} \caption{Results on the JANUS CS2 dataset. The TAR of
all the approaches at FAR=0.1 and 0.01 for the ROC curves. The
Rank-1, Rank-5, and Rank-10 retrieval accuracies of the CMC curves.
We report average and standard deviation of the 10 splits. We also
show the results before and after media averaging where $m$ means
media averaging.} \label{exp:roc_cmc_scores_media}
\end{table*}

\begin{table*}[htp!]
\centering \scriptsize
\begin{tabular}{|c|c|c|c|c|c|c|}
  \hline
  IJB-A-Verif& \cite{wang_face_2015} & JanusB \cite{nist_ijba_2016} & JanusD \cite{nist_ijba_2016} & DCNN$_{bl}$ \cite{aruni_pose_2016}& NAN \cite{yang_neural_2016} & DCNN$_{3d}$ \cite{masi_neural_2016}\\
  \hline
  FAR=1e-3 & 0.514 $\pm$ 0.006& 0.65  & 0.49 & - & 0.785 $\pm$ 0.028&  0.725 \\
  FAR=1e-2 & 0.732 $\pm$ 0.033& 0.826 & 0.71 & - & 0.897 $\pm$ 0.01 &  0.886 \\
  FAR=1e-1 & 0.895 $\pm$ 0.013& 0.932 & 0.89 & - & 0.959 $\pm$ 0.005&  -\\
  \hline\hline
  IJB-A-Ident & \cite{wang_face_2015} & JanusB \cite{nist_ijba_2016} & JanusD \cite{nist_ijba_2016} & DCNN$_{bl}$ \cite{aruni_pose_2016}& NAN \cite{yang_neural_2016} & DCNN$_{3d}$ \cite{masi_neural_2016}\\
  \hline
  Rank-1  & 0.820 $\pm$ 0.024& 0.87  & 0.88  & 0.895 $\pm$ 0.011 & - & 0.906\\
  Rank-5  & 0.929 $\pm$ 0.013& - & - & 0.963 $\pm$ 0.005 & - & 0.962\\
  Rank-10 & - & 0.95  & 0.97  & - & - & 0.977 \\
  \hline
  \hline
  IJB-A-Verif & DCNN$_{pose}$ \cite{AbdAlmageed_pose_2016} & DCNN$_{m}$ (setup 1) & DCNN$_{m}$ (setup 2) & DCNN$_{m}$ (setup 3)& DCNN${tpe}$ \cite{swami_btas_2016}& TP \cite{crosswhite_template_2016} \\
  \hline
  FAR=1e-3 &  -    &  0.704 $\pm$ 0.037 & 0.762 $\pm$ 0.038 & 0.76 $\pm$ 0.038  & \textbf{0.813 $\pm$ 0.02}  &   - \\
  FAR=1e-2 & 0.787 &  0.863 $\pm$ 0.02  & 0.885 $\pm$ 0.014 & 0.889 $\pm$ 0.016 & 0.9   $\pm$ 0.01  & \textbf{0.939 $\pm$ 0.013}\\
  FAR=1e-1 & 0.911 &  0.966 $\pm$ 0.05  & 0.954 $\pm$ 0.003 & \textbf{0.968 $\pm$ 0.005}& 0.964 $\pm$ 0.01 & - \\
  \hline\hline
  IJB-A-Ident & DCNN$_{pose}$ \cite{AbdAlmageed_pose_2016} & DCNN$_{m}$ (setup 1) & DCNN$_{m}$ (setup 2) & DCNN$_{m}$ (setup 3)& DCNN${tpe}$ \cite{swami_btas_2016}& TP \cite{crosswhite_template_2016} \\
  \hline
  Rank-1  & 0.846 & 0.916 $\pm$ 0.009 & 0.923 $\pm$ 0.01 &\textbf{0.942 $\pm$ 0.008}& 0.932 $\pm$ 0.001  & 0.928 $\pm$ 0.01 \\
  Rank-5  & 0.927 & 0.971 $\pm$ 0.007 & 0.961 $\pm$ 0.006&\textbf{0.98  $\pm$ 0.005}& - & -\\
  Rank-10 & 0.947 & 0.981 $\pm$ 0.005&  0.969 $\pm$ 0.004&\textbf{0.988 $\pm$ 0.003}& 0.977 $\pm$ 0.005  & 0.986 $\pm$ 0.003 \\
  \hline
\end{tabular}
 \vspace{+1mm} \caption{Results on the IJB-A
dataset. The TAR of all the approaches at FAR=0.1, 0.01, and 0.001
for the ROC curves (IJB-A 1:1 verification). The Rank-1, Rank-5, and
Rank-10 retrieval accuracies of the CMC curves (IJB-A 1:N
identfication). We report average and standard deviation of the 10
splits. All the performance results reported in
\cite{nist_ijba_2016}, Janus B (JanusB-092015), Janus D
(JanusD-071715), DCNN$_{bl}$ \cite{aruni_pose_2016}, DCNN$_{3d}$
\cite{masi_neural_2016}, NAN \cite{yang_neural_2016}, DCNN$_{pose}$
\cite{AbdAlmageed_pose_2016}, DCNN$_{tpe}$ \cite{swami_btas_2016},
and TP \cite{crosswhite_template_2016} are included in the Table.
 Some of these systems have produced results for setup 1 (based on landmarks provided along with the dataset) only. In addition, we also compare the performance of the
 recent work, DCNN$_{tpe}$ \cite{swami_btas_2016} where the performance difference mainly comes from the better preprocessing module and improved metric, \cite{hyperface}.}
\label{exp:roc_cmc_scores_ijba}
\end{table*}

%\end{savenotes}
Besides using the average feature representation, we also perform
media averaging which is to first average the features coming from
the same media (image or video) and then further average the media
average features to generate the final feature representation. We
show the results before and after media averaging for both IJB-A and
JANUS CS2 dataset in Table~\ref{exp:roc_cmc_scores_ijba_media} and
in Table~\ref{exp:roc_cmc_scores_media} respectively. It is clear
that media averaging significantly improves the performance.

Tables~\ref{exp:roc_cmc_scores_ijba} and~\ref{exp:roc_cmc_scores}
summarize the scores (\emph{i.e.}, both ROC and CMC numbers)
produced by different face verification methods on the IJB-A and
JANUS CS2 datasets, respectively. For the IJB-A dataset, we compare
our fusion results (\emph{i.e.}, we perform finetuning and TSE in
Setup 3.) with DCNN$_{bl}$ (bilinear CNN \cite{aruni_pose_2016}),
DCNN$_{pose}$ (multi-pose DCNN models \cite{AbdAlmageed_pose_2016}),
NAN~\cite{yang_neural_2016}, DCNN$_{3d}$ \cite{masi_neural_2016},
template adaptation (TP) \cite{crosswhite_template_2016},
DCNN$_{tpe}$ \cite{swami_btas_2016} and the ones
\cite{nist_ijba_2016} reported recently by NIST where JanusB-092015
achieved the best verification results, and JanusD-071715 the best
identification results. For the JANUS CS2 dataset,
Table~\ref{exp:roc_cmc_scores} includes, a DCNN-based method
\cite{wang_face_2015}, Fisher vector-based method
\cite{simonyan_fisher_2013}, DCNN$_{pose}$
\cite{AbdAlmageed_pose_2016}, DCNN$_{3d}$ \cite{masi_neural_2016},
and two commercial off-the-shelf matchers, COTS and GOTS
\cite{klare_janus_2015}.
% Figure~\ref{exp:cs2_ijba_roc_fusion} shows the ROC curves
%corresponding to different methods on the JANUS CS2 and IJB-A
% datasets, respectively, for the verification.
From the ROC and CMC scores, we see that the fusion of DCNN methods
significantly improve the performance. This can be attributed to the
fact that the DCNN model does capture face variations over a large
dataset and generalizes well to a new small dataset. In addition,
the performance results of Janus B (Jan-usB-092015), Janus D
(JanusD-071715), DCNN$_{bl}$ and DCNN$_{pose}$ systems have produced
results for setup 1 (based on landmarks provided along with the
dataset) only.

\begin{table*}[htp!]
\centering
\begin{tabular}{|c|c|c|c|c|}
  \hline
  CS2-Verif& COTS & GOTS &FV\cite{simonyan_fisher_2013}& DCNN$_{pose}$\cite{AbdAlmageed_pose_2016}\\
  \hline
  FAR=1e-3&      -        &          -      &       -         &   -    \\
  FAR=1e-2&0.581$\pm$0.054& 0.467$\pm$0.066 & 0.411$\pm$0.081 & 0.897  \\
  FAR=1e-1&0.767$\pm$0.015& 0.675$\pm$0.015 & 0.704$\pm$0.028 & 0.959  \\
  \hline
  CS2-Ident& COTS & GOTS &FV\cite{simonyan_fisher_2013}& DCNN$_{pose}$\cite{AbdAlmageed_pose_2016}\\
  \hline
  Rank-1&0.551 $\pm$ 0.003&0.413 $\pm$ 0.022&0.381 $\pm$ 0.018& 0.865 \\
  Rank-5&0.694 $\pm$ 0.017&0.571 $\pm$ 0.017&0.559 $\pm$ 0.021& 0.934 \\
  Rank-10&0.741 $\pm$ 0.017&0.624 $\pm$ 0.018&0.637 $\pm$ 0.025& 0.949 \\
  \hline
  \hline
  CS2-Verif& DCNN$_{3d}$ \cite{masi_neural_2016} & DCNN (setup 1) & DCNN (setup 2) & DCNN (setup 3)\\
  \hline
  FAR=1e-3& 0.824 & 0.81 $\pm$ 0.018  & 0.823 $\pm$ 0.013 & \textbf{0.83 $\pm$ 0.014} \\
  FAR=1e-2& 0.926 & 0.92  $\pm$ 0.01  & 0.922 $\pm$ 0.008 & \textbf{0.935 $\pm$ 0.007}\\
  FAR=1e-1& - & 0.981 $\pm$ 0.003    & 0.968 $\pm$ 0.003 & \textbf{0.986 $\pm$ 0.002}\\
  \hline
  CS2-Ident& DCNN$_{3d}$ \cite{masi_neural_2016} & DCNN (setup 1) & DCNN (setup 2) & DCNN (setup 3)\\
  \hline
  Rank-1&  0.898 & 0.905 $\pm$ 0.007   & 0.915 $\pm$ 0.007& \textbf{0.931 $\pm$ 0.007}\\
  Rank-5&  0.956 & 0.965 $\pm$ 0.004& 0.959 $\pm$ 0.005& \textbf{0.976 $\pm$ 0.004}\\
  Rank-10& 0.969 & 0.977 $\pm$ 0.004 & 0.967 $\pm$ 0.004& \textbf{0.985 $\pm$ 0.002}\\
  \hline
\end{tabular}
\vspace{+1mm} \caption{Results on the JANUS CS2 dataset. The TAR of
all the approaches at FAR=0.1, 0.01, and 0.001 for the ROC curves.
The Rank-1, Rank-5, and Rank-10 retrieval accuracies of the CMC
curves. We report average and standard deviation of the 10 splits.
The performance results of DCNN$_{pose}$ have produced results for
setup 1 only.} \label{exp:roc_cmc_scores}
\end{table*}

During the review period of the paper, newer results on IJB-A
datasets have been reported. The interested readers are referred
to~\cite{xiong2017good,ranjan2017l2} for more details. In addition,
the NAN~\cite{yang2017neural} results are based on an earlier
version~\cite{yang_neural_2016}. More recent state of the art
results are reported in~\cite{ranjan2017l2} obtained by employing
the deep residual network and $L_2$-norm regularized softmax loss.

\subsection{Labeled Faces in the Wild}
We also evaluate our approach on the well-known LFW dataset
\cite{huang_labeled_2008} using the standard protocol which defines
3,000 positive pairs and 3,000 negative pairs in total and further
splits them into 10 disjoint subsets for cross validation. Each
subset contains 300 positive and 300 negative pairs. It contains
7,701 images of 4,281 subjects. We compare the mean accuracy of the
proposed deep model with other state-of-the-art deep learning-based
methods: DeepFace \cite{taigman_deepface_2014}, DeepID2
\cite{sun_deeply_2014}, DeepID3 \cite{sun_deepid3_2015}, FaceNet
\cite{schroff_facenet_2015}, Yi \emph{et al.}
\cite{yi_learning_2014}, Wang \emph{et al.} \cite{wang_face_2015},
Ding \emph{et al.} \cite{ding_robust_2015}, Parkhi \emph{et al.}
\cite{parkhi_deep_2015}, and human performance on the ``funneled"
LFW images.  The results are summarized in Table ~\ref{exp:acc_lfw}.
It can be seen that our approach performs comparable to other deep
learning-based methods. Note that some of the deep learning-based
methods compared in Table~\ref{exp:acc_lfw} use millions of data
samples for training the model.  In comparison, we use only the
CASIA dataset for training our model which has less than 500K
images.

\begin{table*}[htp!]
\centering \scriptsize
\begin{tabular}{|l|c|l|l|l|}
  \hline
  Method          &\#Net& Training Set & Metric & Mean Accuracy $\pm$ Std\\
  \hline\hline
  DeepFace \cite{taigman_deepface_2014}&  1  & 4.4 million images of 4,030 subjects, private     & cosine                    & 95.92\% $\pm$ 0.29\% \\
  DeepFace                             &  7  & 4.4 million images of 4,030 subjects, private     & unrestricted, SVM         & 97.35\% $\pm$ 0.25\% \\
  DeepID2 \cite{sun_deeply_2014}       &  1  & 202,595 images of 10,117 subjects, private        & unrestricted, Joint-Bayes & 95.43\% \\
  DeepID2                              & 25  & 202,595 images of 10,117 subjects, private        & unrestricted, Joint-Bayes & 99.15\% $\pm$ 0.15\% \\
  DeepID3 \cite{sun_deepid3_2015}      & 50  & 202,595 images of 10,117 subjects, private        & unrestricted, Joint-Bayes & 99.53\% $\pm$ 0.10\% \\
  FaceNet \cite{schroff_facenet_2015}  &  1  & 260 million images of 8 million subjects, private & L2                        & 99.63\% $\pm$ 0.09\% \\
  Yi \emph{et al.} \cite{yi_learning_2014}    &  1  & 494,414 images of 10,575 subjects, public         & cosine                    & 96.13\% $\pm$ 0.30\% \\
  Yi \emph{et al.}                            &  1  & 494,414 images of 10,575 subjects, public         & unrestricted, Joint-Bayes & 97.73\% $\pm$ 0.31\% \\
  Wang \emph{et al.} \cite{wang_face_2015}    &  1  & 494,414 images of 10,575 subjects, public         & cosine                    & 96.95\% $\pm$ 1.02\% \\
  Wang \emph{et al.}                          &  7  & 494,414 images of 10,575 subjects, public         & cosine                    & 97.52\% $\pm$ 0.76\% \\
  Wang \emph{et al.}                          &  1  & 494,414 images of 10,575 subjects, public         & unrestricted, Joint-Bayes & 97.45\% $\pm$ 0.99\% \\
  Wang \emph{et al.}                          &  7  & 494,414 images of 10,575 subjects, public         & unrestricted, Joint-Bayes & 98.23\% $\pm$ 0.68\% \\
  Ding \emph{et al.} \cite{ding_robust_2015}  &  8  & 471,592 images of 9,000 subjects, public         & unrestricted, Joint-Bayes & 99.02\% $\pm$ 0.19\% \\
  Parkhi \emph{et al.} \cite{parkhi_deep_2015}  &  1  & 2.6 million images of 2,622 subjects, public         & unrestricted, TDE & 98.95 \% \\
  Human, funneled \cite{wang_face_2015}& N/A & N/A & N/A & 99.20\% \\
  \hline
  Our DCNN$_S$              &  1  & 490,356 images of 10,548 subjects, public         & cosine                    & 97.7\% $\pm$ 0.8\% \\
  Our DCNN$_L$              &  1  & 490,356 images of 10,548 subjects, public         & cosine                    & 96.8\% $\pm$ 0.6\%\\
  Our DCNN$_S$ + DCNN$_L$   &  2  & 490,356 images of 10,548 subjects, public         & cosine                    & 98\% $\pm$ 0.5\% \\
  Our DCNN$_S$ + DCNN$_L$   &  2  & 490,356 images of 10,548 subjects, public         & unrestricted, TSE & 98.33\% $\pm$ 0.7\% \\
  \hline
\end{tabular}
\vspace{+1mm} \caption{Accuracy of different methods on the LFW
dataset.} \label{exp:acc_lfw}
\end{table*}

\subsection{Comparison with Methods based on Annotated Metadata}
Most systems compared in this paper produced the results for setup 1
which is based on landmarks provided along with the dataset only
(\emph{i.e.,} except DCNN$_{tpe}$.). For DCNN$_{3d}$
\cite{masi_neural_2016}, the number of face images is augmented
along with the original CASIA-WebFace dataset by aro-und 2 million
using 3D morphable models. On the other hand, NAN
\cite{yang_neural_2016} and TP \cite{crosswhite_template_2016} used
datasets with more than 2 million face images to train the model.
However, the networks used in this work were trained with the
original CASIA-WebFace which contains around 500K images. In
addition, TP adapted the one-shot similarity framework
\cite{wolf2009one} with linear support vector machine for set-based
face verification and trained the metric on-the-fly with the help of
a pre-selected negative set during testing. Although TP achieved
significantly better results than other approaches, it takes more
time during testing than the proposed method since our metric is
trained off-line and requires much less time for testing than TP. We
expect the performance of the proposed approach can also be improved
by using the one-shot similarity framework. As shown in
Table~\ref{exp:roc_cmc_scores_ijba}, the proposed approach achieves
comparable results to other methods and strikes a balance between
testing time and performance. In a recent work, DCNN$_{tpe}$
\cite{swami_btas_2016}, adopted a probabilistic embedding for
similarity computation and a new face preprocessing module,
hyperface \cite{hyperface}, for improved face detection and
fiducials where \cite{hyperface} is a multi-task deep network
trained for the tasks of gender classification, fiducial detection,
pose estimation and face detection. We plan to incorporate hyperface
into the current framework which may yield some improvement in
performance.

\subsection{Run Time}
The DCNN$_{S}$ model for face verification is trained on the
CASIA-Webface dataset from scratch for about 4 days and for
DCNN$_{L}$, it takes 20 hours to train on the same face dataset
which is initialized using the weights of Alexnet pretrained on the
ImageNet dataset. The two networks are trained using NVidia Titan X
with cudnn v4. The running time for face detection is around 0.7
second per image. The facial landmark detection and feature
extraction steps take about 1 second and 0.006 second per face,
respectively (\emph{i.e.}, To compare the speed difference, we run the feature extraction part using CPU.
it takes around 0.7 second for feature extraction using a core of 16-core 3.0GHz Intel Xeon CPU and math library atlas
which is around 100 times as the GPU time.) The face association module for a video takes around 5
fps on average.

\section{Open Issues}\label{sec:open}
Given sufficient number of annotated data and GPUs, DCNNs have been
shown to yield impressive performance improvements.  Still many
issues remain to be addressed to make the DCNN-based recognition
systems robust and practical. These are briefly discussed below.

\begin{itemize}
\item {\textbf{Reliance on large training data sets:}} One of the
  top performing networks in the MegaFace challenge needs 500 million faces of about 10 million subjects. Such
  large annotated training set may not be always available (e.g. expression
  recognition, age estimation). So networks that can perform well with
  reasonable-sized training data are needed.

\item {\textbf{Invariance:}} While limited invariance to translation is possible
  with existing DCNNs, networks that can incorporate more general invariances
  are needed.

\item {\textbf{Training time:}} The training time even when GPUs are used can be
  several tens of hours, depending on the number of layers used and
  the training data size. More efficient implementations of learning algorithms,
  preferably implemented using CPUs are desired.

\item {\textbf{Number of parameters:}}  The number of parameters can be several
  tens of millions. Novel strategies that reduce the number of parameters need
  to be developed.

\item {\textbf{Handling degradations in training data:}} : DCNNs robust to
  low-resolution, blur, illumination and pose variations, occlusion, erroneous annotation,  etc. are needed to
  handle degradations in data.

\item {\textbf{Domain adaptation of DCNNs:}} While having large volumes of data
  may help with processing test data from a different distribution than that of
  the training data, systematic methods for adapting the deep features to test
  data are needed.

\item {\textbf{Theoretical considerations:}} While DCNNs have been around for a
  few years, detailed theoretical understanding is just starting to develop \cite{bruna_invariant_2013,mallat_understanding_2016,Raja_deep,vidal_deep}.
  Methods for deciding the number of layers, neighborhoods over which max pooling operations
  are performed are needed.

\item {\textbf{Incorporating domain knowledge:}} The current practice is to rely
  on fine tuning. For example, for the age estimation problem, one can start
  with one of the standard networks such as the AlexNet and fine tune it using
  aging data. While this may be reasonable for somewhat related problems (face
  recognition and facial expression recognition), such fine tuning strategies
  may not always be effective. Methods that can incorporate context may make the
  DCNNs more applicable to a wider variety of problems.

\item {\textbf{Memory:}} Although Recurrent CNNs are on the rise, they
  still consume a lot of time and memory for training and deployment.
  Efficient DCNN algorithms are needed to handle videos and
  other data streams as blocks.

\end{itemize}

We also discussed design considerations for each component of a full
face verification system, including

\begin{itemize}
\item {\textbf{Face detection:}} Face detection is challenging due to the wide range of variations in the appearance of faces. The variability is caused mainly by changes in illumination, facial expression, viewpoints, occlusions, etc.
Other factors such as blurry images and low resolution are prominent
in face detection task.

\item {\textbf{Fiducial detection:}} Most of the datasets only contain few thousands images.
A large scale annotated and unconstrained dataset will make the face
alignment system more robust to the challenges, including extreme
pose, low illumination, small and blurry face images. Researchers
have hypothesized that deeper layers can encode more abstract
information such as identity, pose, and attributes; However, it has
not yet been thoroughly studied which layers exactly correspond to
local features for fiducial detection.

\item {\textbf{Face association:}}
Since the video clips may contain media of low-quality images, the
blurred and low-resolution image makes the face detection not
reliable. This may lead to performance degradation of face
association since a face track will not be initiated due to the
missing of face detection. Besides, abrupt motion, occlusion, and
crowded scene can lead to performance degradation of tracking and
potential identity switching.

\item {\textbf{Face verification:}} For face verification, the performance can be improved by learning a discriminative
distance measure. However, due to memory constraints limited by
graphics cards, how to choose informative pairs or triplets and
train the network end-to-end using online methods (\emph{e.g.},
stochastic gradient descent) on large-scale datasets are still open
problems.
\end{itemize}

\section{Conclusion} \label{sec:conc}
We presented the design and performance of our automatic face
verification system, which automatically locates faces and performs
verification/recognition on newly released challenging face
verification datasets, IARPA Benchmark A (IJB-A) and its extended
version, JANUS CS2. It is shown that the proposed DCNN-based system
can not only accurately locate the faces across images and videos
but also learn a robust model for face verification. Experimental
results demonstrate that the performance of the proposed system on
the IJB-A dataset is much better than a FV-based method and some
COTS and GOTS matchers. \\

\section{Acknowledgments}
This research is based upon work supported by the Office of the
Director of National Intelligence (ODNI), Intelligence Advanced
Research Projects Activity (IARPA), via IARPA R\&D Contract No.
2014-14071600012. The views and conclusions contained herein are
those of the authors and should not be interpreted as necessarily
representing the official policies or endorsements, either expressed
or implied, of the ODNI, IARPA, or the U.S. Government. The U.S.
Government is authorized to reproduce and distribute reprints for
Governmental purposes notwithstanding any copyright annotation
thereon. We thank professor Alice O'Toole for carefully reading the
manuscript and suggesting improvements in the presentation of this
work.

%\begin{acknowledgements}
%\end{acknowledgements}

\bibliographystyle{spmpsci}      % mathematics and physical sciences
\bibliography{all}   % name your BibTeX data base

%\begin{thebibliography}{}
%%
%% and use \bibitem to create references. Consult the Instructions
%% for authors for reference list style.
%\end{thebibliography}

\end{document}